\documentclass[10pt,twocolumn,letterpaper]{article}

\usepackage{cvpr}              

\usepackage{amsmath}
\usepackage{amssymb}
\usepackage{bm}
\usepackage{wrapfig} 
\usepackage{pifont}
\usepackage{threeparttable}
\usepackage{framed}
\usepackage{makecell}
\usepackage{amsfonts}
\usepackage{algorithmic}
\usepackage{graphicx}
\usepackage{textcomp}
\usepackage{arydshln}
\usepackage{booktabs}
\usepackage{multirow}
\usepackage{array}
\usepackage{lipsum} 
\usepackage{enumitem}
\usepackage[export]{adjustbox}
\usepackage{dblfloatfix}
\usepackage{tabulary}
\usepackage{siunitx}
\usepackage{bbm}
\usepackage{algorithm}
\usepackage{algorithmic}
\usepackage{listings}
\usepackage{xcolor}

\definecolor{delay}{RGB}{230,94,42}
\definecolor{mywarning}{RGB}{233,144,61}
\definecolor{mygray}{gray}{.9}
\definecolor{mygray2}{gray}{.52}
\definecolor{ggray}{RGB}{127,127,127}
\definecolor{reda}{RGB}{192,0,0}
\definecolor{redb}{RGB}{217,148,143}
\definecolor{myyellow}{RGB}{190,144,0}
\definecolor{mygreen}{RGB}{80,100,40}
\definecolor{myblue}{RGB}{30,90,100}
\definecolor{codegreen}{RGB}{79,126,127}
\definecolor{codedefine}{RGB}{153,54,159}
\definecolor{codefunc}{RGB}{73,122,234}
\definecolor{codecall}{RGB}{73,122,234}
\definecolor{codepro}{RGB}{212,96,80}
\definecolor{codedim}{RGB}{89,152,195}
\usepackage[OT1]{fontenc} 

\usepackage[pagebackref,breaklinks,colorlinks,allcolors=cvprblue]{hyperref}

\usepackage[capitalize]{cleveref}
\crefname{section}{Sec.}{Secs.}
\Crefname{section}{Section}{Sections}
\Crefname{table}{Table}{Tables}
\crefname{table}{Tab.}{Tabs.}

\definecolor{graymy}{rgb}{0.5,0.5,0.5}
\definecolor{mygray}{gray}{.9}
\definecolor{cvprblue}{rgb}{0.21,0.49,0.74}
\makeatletter
\newcommand{\thickhline}{%
	\noalign {\ifnum 0=`}\fi \hrule height 1pt
	\futurelet \reserved@a \@xhline
}
\makeatother

\usepackage{lipsum}

\def\paperID{7901} 
\def\confName{CVPR}
\def\confYear{2025}

\author{
Mu Chen\textsuperscript{1}, Liulei Li\textsuperscript{2}, Wenguan Wang\textsuperscript{1}\footnotemark[2],~~Yi Yang\textsuperscript{1}\\
\small \textsuperscript{1} ReLER, CCAI, Zhejiang University~~\textsuperscript{2} ReLER, AAII, University of Technology Sydney \\
\url{https://github.com/kagawa588/DiffVsgg}\\
\small\url{}
}

\begin{document}

\title{\textsc{DiffVsgg}: Diffusion-Driven Online Video Scene Graph Generation\\ }

\maketitle

\begin{abstract}
\footnotetext[2]{Corresponding author: Wenguan Wang.}
Top-leading solutions for Video Scene Graph Generation (VSGG)
typically adopt an offline pipeline.
Though demonstrating promising performance, they remain unable to handle real-time video streams and consume large GPU memory.  Moreover, these approaches fall short in temporal reasoning, merely aggregating frame-level predictions over a temporal context. 
In response, we introduce \textsc{DiffVsgg}, an online VSGG solution that frames this task as an iterative scene graph update problem.  Drawing inspiration from Latent Diffusion Models (LDMs) which generate images via denoising a latent feature embedding, we unify the decoding of object classification, bounding box regression, and graph generation three tasks using one shared feature embedding.
Then, given an embedding containing unified features of object pairs, we conduct a step-wise Denoising on it within LDMs, so as to deliver a clean embedding which clearly indicates the relationships between objects.
This embedding then serves as the input to task-specific heads for object classification, scene graph generation, etc.
\textsc{DiffVsgg} further facilitates continuous temporal reasoning, where predictions for subsequent frames leverage results of past frames as the conditional inputs of LDMs, to guide the reverse diffusion process for current frames.
Extensive experiments on three setups of Action Genome demonstrate the superiority of \textsc{DiffVsgg}. 
\end{abstract}    
    

\section{Introduction} \label{sec:intro}

Video Scene Graph Generation (VSGG)  is receiving growing attention as it benefits a wide range of downstream tasks (\eg, video caption~\cite{guadarrama2013youtube2text,venugopalan2015sequence}, video retrieval~\cite{gao2017tall,lei2020tvr}, and visual question answering ~\cite{lei2018tvqa,tapaswi2016movieqa}). 
To deliver a holistic understanding of the underlying spatial-temporal dynamics within scenes, this task aims to construct a sequence of directed graphs where nodes represent objects and edges describe inter-object relationships (\textit{a.k.a.}, \textit{predicate}). 
 
   \begin{figure}[t]
      \begin{center}
          \includegraphics[width=1.\linewidth]{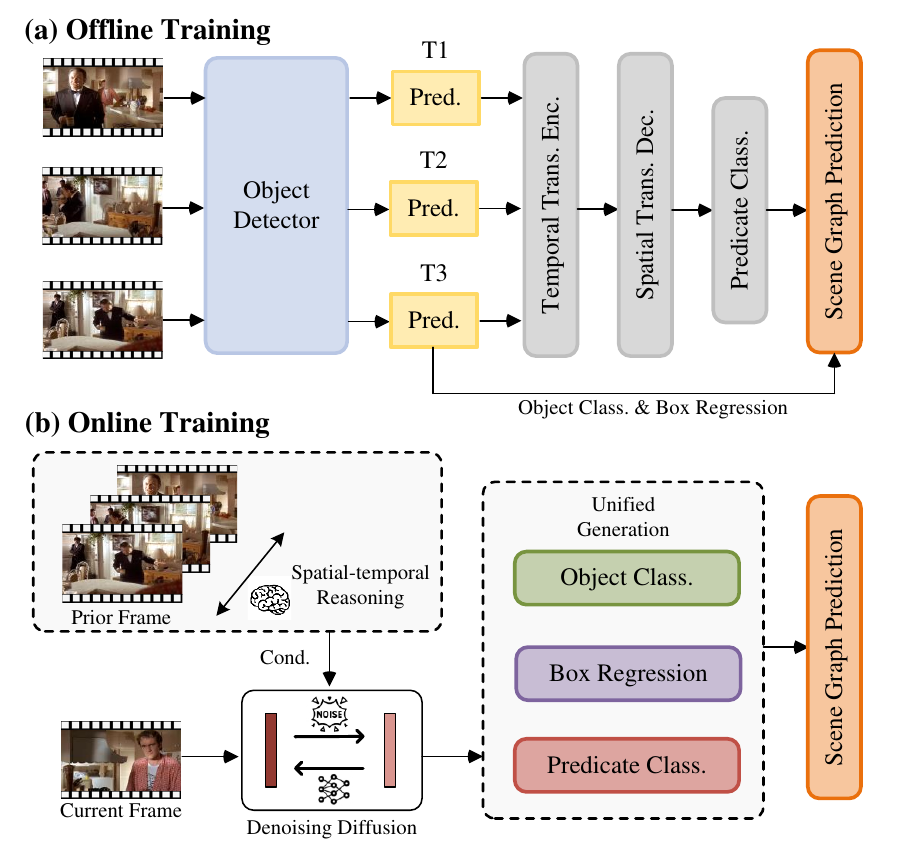}
          \end{center}
      \vspace{-11pt}
      \captionsetup{font=small}
      \caption{\small{(a) Existing VSGG solutions typically adopt an offline training pipeline, dividing the problem into various components: object detection, temporal association, and contextual aggregation. (b) \textsc{DiffVsgg} introduces a new paradigm that performs spatial-temporal reasoning directly as each frame is processed sequentially, enabling progressive, online updates to the scene graph.
      }
      }
      \label{fig:1}
      \vspace{-7pt}
    \end{figure}

Top-leading VSGG approaches typically adopt an offline pipeline\!~\cite{feng2023exploiting,wang2023cross,nag2023unbiased}, 
where scene graphs are generated independently for each frame and then aggregated along the temporal dimension (\ie, Fig.\!~\ref{fig:1}(a)).
Though demonstrating strong performance, 
they require full video sequences as inputs, which faces challenges in processing long videos 
containing hundreds of frames 
due to GPU memory constraints, and is unable to handle real-time video streams for applications like autonomous driving and augmented reality\!~\cite{wu2022defense}. 
Moreover, the dealing of temporal cues focuses solely on the global aggregation of frame-level predictions 
via Transform blocks.
This falls short in the reasoning over temporal space, which is essential for modeling the dynamic changes of interactions between subjects and objects, and potentially benefits predicate prediction.
Though a surge of early work has sought to facilitate explicit reasoning over the temporal domain via messaging passing\!~\cite{tsai2019video} or spatial-temporal graph\!~\cite{qian2019video},  a large performance gap remains when compared to these offline temporal aggregation approaches.

To address  these challenges, we propose \textsc{DiffVsgg}, a high performance online approach for VSGG in leverage of Latent Diffusion Models (LDMs)\!~\cite{sohl2015deep,rombach2022high} (\ie, Fig.\!~\ref{fig:1}(b)). 
In VSGG, the scene graph is dynamically evolved throughout the progression of video frames, with nodes and edges being continuously updated to precisely reflect the latest video content.
Such an iterative update process shares a similar spirit of LDMs, which progressively remove noise to generate new samples from data distributions.
Naturally, our motivation is to develop a VSGG model that aligns with the denoising principle of LDMs to iterative refine scene graphs along the temporal dimension, facilitating online temporal reasoning within this reverse diffusion process.

To achieve this, we tackle VSGG from a unified perspective, organizing the decoding of object classification, bounding box regression, and graph generation three tasks from a shared feature embedding, which serves as the input and output of the Denoising U-Net. Concretely, from the \textit{spatial perspective}, given the object detection results from each frame, multiple object pairs can be constructed. Then, for each object pair, we integrate \textbf{i)} the visual feature of two objects, \textbf{ii)} the union feature between, and \textbf{iii)} the locations of two objects, into a unified embedding. This embedding serves as the input to the denoising U-Net, after steps of denoising, a clean embedding clearly describing the objects as well as their inter-object relationships can be delivered.  
From the \textit{temporal perspective}, \textsc{DiffVsgg} conducts frame-by-frame reasoning where the result for each subsequent frame is delivered by iteratively refining the predictions of previous frames via reverse diffusion. Specifically, object positions and contextual information from prior frames are leveraged as conditions to guide the denoising of the shared feature embeddings for current frames.
This encourages continuous temporal reasoning as the video progresses, allowing \textsc{DiffVsgg} to effectively capture long-term spatiotemporal dependencies and adapt to complex motion patterns in an online manner.
To further unlock the potential of LDMs, we build a memory bank to store positions for each object. In this way, motion information including acceleration and deceleration of object can be explicitly calculated. These motion cues are integrated into conditional inputs for the reverse diffusion step, which can help infer relationships such as \texttt{following} or \texttt{approaching} between objects.

\textsc{DiffVsgg} distinguishes itself in several aspects: 
\textbf{First}, it tackles VSGG in an online manner to continuously address an unlimited number of frames, while being friendly to devices with limited GPU memory. 
\textbf{Second}, the prediction of each video is implemented as a reverse diffusion process, which elegantly encodes spatial and temporal reasoning into the Denoising step.
\textbf{Third}, the temporal cues (\ie, historic predictions) are propagated into subsequent frames as conditions and participating in the prediction, while prior work simply aggregates predictions of all frames along the temporal dimension.
\textbf{Fourth}, the learning of object classification, bounding box regression, and graph generation three tasks are  jointly optimized with a shared feature embedding.
Such a unified learning paradigm allows for the solving of VSGG from a global view, 
and avoids error made in one task propagating to subsequent tasks.
\textbf{Fifth}, this unified decoding simplifies the VSGG pipeline and eliminates cumbersome handcrafted modules such as non-maximum suppression (NMS), and entity matching across frames.

To the best of our knowledge, \textsc{DiffVsgg} is the first work that treats the VSGG task as an iterative denoising problem along the temporal dimension. Extensive experiments on Action Genome (AG) \!~\cite{ji2020action} demonstrate the superiority of our proposed method. Notably,$_{\!}$ \textsc{DiffVsgg}$_{\!}$ achieves$_{\!}$ SOTA$_{\!}$ results across all three setups and surpasses the top-leading solutions (\eg, DSG-DETR\!~\cite{feng2023exploiting}) by \textbf{3.3} in terms of R@10,
highlighting the great potential of utilizing diffusion models for visual relation understanding in videos.

\section{Related Work} \label{sec:relatedwork}

\noindent\textbf{Scene Graph Generation (SGG).} 
SGG involves detecting object instances and classifying their pairwise visual relations in an image, which is essential for comprehensive visual tasks~\cite{wang2025visual,wei2024nonverbal,zhou2020cascaded,zhou2021cascaded,lineural}. Recent SGG approaches seek to comprehend visual context by aggregating spatial context through various strategies, including explicit message passing\!~\cite{xu2017scene,li2017vip,zellers2018neural,chen2019knowledge,li2017scene,dai2017detecting}, graph structure modeling\!~\cite{tang2019learning,yin2018zoom,yang2018graph,suhail2021energy,woo2018linknet}, external knowledge integration\!~\cite{chen2019knowledge,gu2019scene,yu2017visual,baier2017improving,yao2021visual,chen2025scene}, and transformer-based networks\!~\cite{cong2023reltr,li2022sgtr,shit2022relationformer,liang2019vrr,jung2023devil,lu2021context,qi2019attentive,dong2022stacked,dong2021visual,kundu2023ggt,chen2025iclr}.

Extended from SGG, Video Scene Graph Generation (VSGG) aims to ground visual relationships jointly in spatial and temporal dimensions.  Prior research primarily concentrates on addressing the long-tail distribution problem observed in prominent benchmarks~\cite{ji2020action}. Numerous unbiased approaches\!~\cite{misra2016seeing, tang2020unbiased, desai2021learning, knyazev2020graph, yan2020pcpl, chiou2021recovering, li2022devil, li2024nicest} have been devised to handle  infrequent predicate classes arising from this distribution.  Furthermore, recent VSGG methods underscore the importance of spatial-temporal learning, and make use of the sequence-processing ability of Transformer\!~\cite{arnab2021vivit, nawhal2021activity, sun2019videobert} to capture temporal continuity. Nonetheless, these approaches often depend on complex post-processing that decouples spatial and temporal learning into two independent steps,  and simply emphasizes on the consistent matching between frames.

In contrast, \textsc{DiffVsgg} handles both spatial and temporal reasoning from a unified perspective, where the reasoning on temporal dimension is naturally achieved via utilizing the spatial reasoning results of prior frames as the conditions to guide predictions for the current frames.

\noindent\textbf{Diffusion Models Beyond Image Generation.}$_{\!}$ 
Diffusion Models (DMs) have surpassed many other generative models in image generation\!~\cite{dhariwaldiffusion,ho2020denoising,ho2022video,cao2024survey,xu2023geometric,vignacdigress} by successively denoising images. Beyond image generation, DMs inherently perform implicit discriminative reasoning while generating data, which proves highly effective in visual tasks that require complex relationship modeling and spatiotemporal reasoning. Therefore, a surge of work has successfully adapted generative diffusion models for tasks including image segmentation\!~\cite{amit2021segdiff,baranchuklabel,brempong2022denoising,chen2023generalist,graikos2022diffusion,kim2023diffusion,wang2023dformer}, object detection\!~\cite{ranasinghe2024monodiff,xu20243difftection,chen2023diffusiondet}, object tracking\!~\cite{luo2024diffusiontrack,lv2024diffmot,xie2024diffusiontrack}, and monocular depth estimation\!~\cite{duan2023diffusiondepth,saxena2023monocular,zhao2023unleashing,mao2024stealing}.   
 Recent research has also utilizes DMs for more complex tasks that require high-level visual understanding abilities such as visual-linguistic understanding\!~\cite{krojer2023diffusion}, scene generation\!~\cite{hong2024human}, and human-object interaction detection\!~\cite{lihuman,jiang2024record}. The potential of DMs in processing graph data has also been explored, encompassing a range of advanced tasks. This includes early work employing score-based methods for generating permutation-invariant graphs\!~\cite{niu2020permutation,jo2022score}, as well as recent approaches focused on enhancing graph neural networks\!~\cite{zhaoflexible,zhaoadaptive} and graph generation techniques\!~\cite{yang2024graphusion}.

In this work, we borrow the step-wise denoising ability of latent diffusion models to enable spatial reasoning within a single frame. This is achieved by iteratively refining the union embeddings of object pairs, which serve to inform the prediction for bounding boxes and predicate classes, \etc.
On the other hand,  temporal reasoning across frames is facilitated  via conditional prompting, \ie, using predictions of prior frames which contain rich location and inter-object relationship cues to guide the denoising of current frames.

\noindent\textbf{Temporal Reasoning in Videos.}$_{\!}$ 
Reasoning within the temporal dimension poses significant challenges for achieving high-level comprehension in video-related tasks~\cite{liu2023bird,fan2019understanding,qi2018learning,yangdoraemongpt,chen2024general}. Recent advancements in video question answering (VQA) seek to tackle temporal reasoning by employing attention mechanisms\!~\cite{fan2019heterogeneous,su2018learning} or incorporating external memory modules\!~\cite{kim2020modality,rahman2021improved} across video frames. In addition, to accurately capture the start, progression, and end of an action,  action recognition approaches\!~\cite{wang2021tdn,truong2022direcformer} model complex motion patterns and dependencies in sequence. Video object detection (VOD)\!~\cite{cui2023feature,liu2023objects} focuses on accurately capturing object trajectories, even in the presence of occlusions and abrupt movements. The top-leading VSGG solutions\!~\cite{feng2023exploiting,wang2023cross,nag2023unbiased} instead aggregate predictions across frames to maintain the temporal consistency of objects. 

In summary, although existing work facilitates temporal reasoning from different perspectives and demonstrates it effectiveness, the potential of diffusion models remains largely$_{\!}$ underexplored.$_{\!}$ In$_{\!}$ fact,$_{\!}$ the sequential denoising in time, where each step is informed by previous one, offers a suitable tool to reconstruct the states for current frames based on observations from past frames. This insight motivates the proposal of \textsc{DiffVsgg}, which tackles the VSGG task via graph denoising over the temporal dimension.

    \begin{figure*}[t]
        \begin{center}
            \includegraphics[width=1\linewidth]{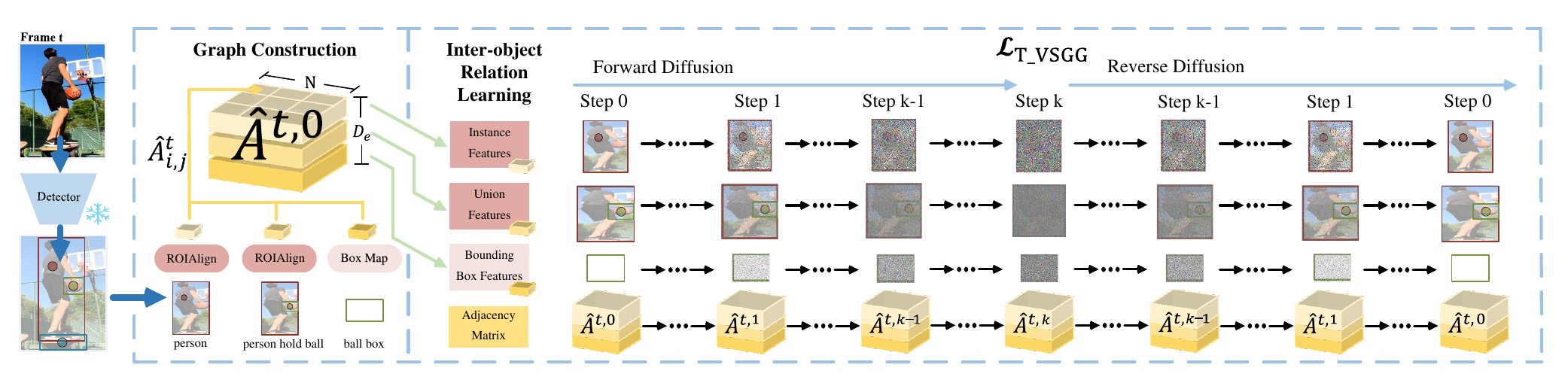}
            \end{center}
        \vspace{-20pt} 
        \captionsetup{font=small}
        \caption{\small{Overview of our proposed inter-object relationship learning strategy using latent diffusion models.}
        }
        \label{fig:2} 
        \vspace{-7pt}
    \end{figure*}

\section{Methodology} \label{sec:methodology}

In this section, we first give a brief introduction to the general background of latent Diffusion Models (\S\ref{sec:preliminary}), then elaborate on the overall design of our proposed \textsc{DiffVsgg} (\S\ref{sec:diffusion}), and finally present the detailed information on the network architecture and training objectives (\S\ref{sec:implementation}).

\subsection{Preliminary: Latent Diffusion Models} \label{sec:preliminary}

To begin, we give an illustration on how diffusion processes\!~\cite{sohl2015deep} are used to model data distributions and generate high-quality samples. Specifically, we consider a continuous-time Markov chain with $t\in [0, T]$:
\vspace{-3pt}  
\begin{equation}
  q(x_t|x_{t-1})=\mathcal{N}(x_t;\sqrt[]{1-\beta_t}x_{t-1}, \beta_t\bm{I}),
  \label{eq:1}
\vspace{-2pt} 
\end{equation}
where $\beta_t$ is a small noise variance term at timestep $t$, and $\mathcal{N}$ denotes a Gaussian distribution. 
The corrupted data at any intermediate time step $t$ can be derived recursively as:
\vspace{-3pt} 
\begin{equation}
\begin{aligned}
  x_t=\sqrt[]{\overline\alpha_t}x_0+\sqrt{1-\overline{\alpha}}\epsilon,
\end{aligned}\label{eq:2}
  \vspace{-2pt} 
\end{equation}
where $\alpha_t=1-\beta_t$ and $\overline\alpha_t=\prod_{s=1}^t\alpha_s$ is the cumulative product of $\alpha_t$
  over the timesteps, $\epsilon\sim\mathcal{N}(\bm{0}, \bm{I})$ represents the sampled noise. 
The reverse process essentially involves learning to reverse the noise addition at each step. 
The cleaned data at time step $t$ can be written recursively as:
\vspace{-3pt} 
\begin{equation}
\begin{aligned}
  x_{t-1}=\frac{1}{\sqrt[]{\alpha_t}}(x_t-\frac{\beta_t}{\sqrt{1-\overline{\alpha}}}\epsilon_\theta(x_t,t),
\end{aligned}\label{eq:3}
  \vspace{-2pt} 
\end{equation}
where a time-conditioned denoising neural network $\epsilon_\theta(x, t)$ is trained to minimize the mean squared error between the true and predicted noise at each timestep:
\vspace{-3pt}
\begin{equation}
\mathcal{L}_\text{DM} := \mathbb{E}_{x_0, \epsilon,  t}\Big[ \Vert \epsilon - \epsilon_\theta(x_t,t) \Vert_{2}^{2}\Big].
\label{eq:dmloss}
\vspace{-2pt} 
\end{equation}
Building on the diffusion process described above, latent diffusion models (LDMs)\!~\cite{rombach2022high} first encode samples into a low-dimensional latent space $z=\mathcal{E}(x)$ using an encoder $\mathcal{E}$,  and then apply the diffusion process within this compressed space. Moreover, LDMs introduce the conditioning mechanisms which allow for control over the generated output based on additional input $y$ such as text, labels, or images. Consequently, the training objective for LDMs is given as:
\vspace{-3pt}
\begin{equation}
\mathcal{L}_\text{LDM} := \mathbb{E}_{\mathcal{E}(x_0),y,  \epsilon,  t}\Big[ \Vert \epsilon - \epsilon_\theta(\bm{z}_{t},t,c_\theta(y)) \Vert_{2}^{2}\Big],
\label{eq:ldmloss}
\vspace{-2pt}
\end{equation}
where $c_\theta$ is a conditioning model to encode $y$.

\subsection{\textsc{\textbf{DiffVsgg}}: Diffusion Models for VSGG}\label{sec:diffusion}
\textbf{Problem Definition.}
Considering a video sequence represented as $\mathcal{I}=\left\{I^1, \cdots, I^T\right\}$ containing $T$ frames, the objective of VSGG is to generate a sequence of scene graph $\mathcal{G}=\left\{{G}^1, \cdots, {G}^T\right\}$ over time, where each element $\mathcal{G}^t$ represents the corresponding scene graph for frame $I^t$.
Each graph $G^t$ is defined as $G^t=\left(\mathcal{V}^t, \mathcal{E}^t\right)$, with $\mathcal{V}^t$ and $\mathcal{E}^t$ denoting the sets of graph nodes and edges, respectively. Each node $v^t_i \in \mathcal{V}^t$ includes attributes such as category and location for object $i$, while each edge $e^t_{i,j} \in \mathcal{E}^t$ describes the inter-object relationship (\textit{a.k.a.}predicate) between subject $i$ and object $j$. 
In this manner, $\mathcal{G}$ captures all objects (\ie, $\mathcal{V}$), their interactions (\ie, $\mathcal{E}$), and the dynamics as they evolve over time (\ie, from $t=1$ to $t=T$) within the scene.

\noindent\textbf{Graph Construction.} Given a video $\mathcal{I}=\left\{I^1, \cdots, I^T\right\}$, we first employ an on-the-shelf object detector $\mathcal{F}_\text{det}$ for each frame:
\vspace{-3pt}
\begin{equation} 
\{\bm{F}^t,\mathcal{B}^t, \mathcal{O}^t\}= \mathcal{F}_\text{det}(I^t),
\label{eq:det}
\vspace{-2pt}
\end{equation}
where $\bm{F}^t$ is the feature extracted by backbone, $\mathcal{B}^t=\{b^t_1,\cdots,b^t_{N_t}\}$ and $\mathcal{O}^t=\{o^t_1,\cdots,o^t_{N_t}\}$ are bounding box and class
predictions for $N_t$ objects detected from frame $I^t$, respectively. 
This serves to initialize an adjacency matrix $\bm{A}^t\in \mathbb{R}^{N_t \times N_t \times D_e}$, of which each element aims to represent the inter-object relationships between object $i$ and $j$:
\vspace{-3pt}
\begin{equation}
\begin{aligned}
\bm{A}_{i,j}^t=  [\bm{F}^t_{o_i}; \bm{F}^t_{o_i, o_j};\bm{F}^t_{b_i}]\in\mathbb{R}^{D_e},\\
\bm{A}_{j,i}^t=  [\bm{F}^t_{o_j}; \bm{F}^t_{o_j, o_i};\bm{F}^t_{b_j}]\in\mathbb{R}^{D_e}.
\label{eq:edge}
\end{aligned}
\vspace{-2pt}
\end{equation}
\textcolor{black}{Here [] refers to the concatenation of features, implemented as \texttt{torch.cat()}.} $\bm{F}^t_{o_i}=\mathcal{F}_\text{ROI}(\bm{F}^t, b^t_i)$ is the instance-level feature for object $i$ extracted via ROIAlign\!~\cite{he2017mask} (\ie, $\mathcal{F}_\text{ROI}$), $\bm{F}^t_{o_i, o_j}=\mathcal{F}_\text{ROI}(\bm{F}^t, b^t_i\cup b_j^t)$ is the feature mapped from the union box of object $i$ and $j$ (\ie, $b^t_i\cup b_j^t$), and  $\bm{F}^t_{b_i}$ describes bounding box $b^t_i$  using a box-to-feature mapping function identical to that in \cite{zellers2018neural}. 
$\bm{A}_{i,j}^t$ represents the relationship predicted from subject $i$, which differs from $\bm{A}_{j,i}^t$ where $j$ is considered the subject. In this way, the matrix $\bm{A}^t$ encodes both node features of each object and the edge features between any pair of them. \textcolor{black}{Note that such subject-based encoding does not incorporate $\bm{F}{o_j}$ and $\bm{F}{b_j}$ into $\bm{A}_{i,j}$ (\ie, $[\bm{F}_{o_i}; \bm{F}_{b_i}; \bm{F}_{o_i, o_j}; \bm{F}_{o_j}; \bm{F}_{b_j}]$), which hinders the neural network to distinguish between the subject ($i$) and object ($j$).} Additionally, as the number of instances (\ie, $N_t$) typically varies across frames due to the emergence or disappearance of some instances,  we pad $\bm{A}^t$ to a fixed size of $N\times N$ where $N>\max(N_{1},\cdots,N_{T})$, using randomly generated feature embedding from Gaussian distributions. 

\noindent\textbf{Inter-object Relationship Learning via LDMs.} 
Next we aim to facilitate the learning of inter-object relationships within LDMs using ground truth scene graph (\ie, Fig.\!~\ref{fig:2}).
Since we use an off-the-self object detector, the extracted feature $\bm{F}^t$ remains static for each frame. Therefore, with access to the ground truth bounding box annotations, we can compute precise edge features via Eq.\!~\ref{eq:edge} from $\bm{F}^t$. The adjacency matrix $\hat{\bm{A}}^t$ which exactly encodes the inter-object features can also be obtained. 
\textcolor{black}{Here $\hat{\bm{A}}^t$ is the ground truth of inter-object relationships, initialized using features derived from ground-truth bounding boxes, where object pairs with no relations are represented as empty entries.} $\hat{\bm{A}}^t$ instructs the learning of a LDM which is \textcolor{black}{responsible for recovering object features and their relationships from random noise.} 

\noindent$\bullet$\textbf{\ \textit{Forward Process.}} Let $\hat{\bm{A}}^{t,0}$ represent the clean adjacency matrix.
The  noise injection process to progressively perturb this matrix is defined following Eq.\!~\ref{eq:2}, and expressed as:
\vspace{-2pt} 
\begin{equation}
\begin{aligned}
  \hat{\bm{A}}^{t,k}=\sqrt[]{\overline\alpha_t}x\hat{\bm{A}}^{t,0}+\sqrt{1-\overline{\alpha}}\epsilon,
\end{aligned}\label{eq:addnoise}
  \vspace{-4pt} 
\end{equation}
where $\hat{\bm{A}}^{t,k}$ denotes the noisy adjacency matrix at step $k$. 
This enables the model to learn robust feature representations, through the  exposure to degraded versions of $\hat{\bm{A}}^{t,0}$.

\noindent$\bullet$\textbf{\ \textit{Reverse Process.}} To recover the original adjacency matrix $\hat{\bm{A}}^{t,0}$, we employ a denoising U-Net $\epsilon_\theta$ which is trained to iteratively remove noise starting from the initial noisy matrix $\hat{\bm{A}}^{t,K}$.
This process follows the denoising step defined in Eq.\!~\ref{eq:3} and proceeds as:
\vspace{-2pt} 
\begin{equation}
\begin{aligned}
  \hat{\bm{A}}^{t,k-1}=\frac{1}{\sqrt[]{\alpha_t}}(\hat{\bm{A}}^{t,k}-\frac{\beta_t}{\sqrt{1-\overline{\alpha}}}\epsilon_\theta(\hat{\bm{A}}^{t,k},k),
\end{aligned}\label{eq:removenoise}
  \vspace{-2pt} 
\end{equation}
where $\epsilon_\theta$ is optimized to recover meaningful inter-object relationships. Following Eq.\!~\ref{eq:ldmloss}, we define the training objective $\mathcal{L}_\text{VSGG}$ by minimizing the spatial structure loss:
\vspace{-2pt} 
\begin{equation}
\mathcal{L}_\text{VSGG} := \mathbb{E}_{\hat{\bm{A}}^{t,k}, \epsilon,  k}\Big[ \Vert \epsilon - \epsilon_\theta(\hat{\bm{A}}^{t,k},k) \Vert_{2}^{2}\Big].
\label{eq:spatialloss}
\vspace{-2pt}
\end{equation}
$\mathcal{L}_\text{VSGG}$ encourages $\epsilon_\theta$ to accurately predict and remove the noise at each step $k$, so as  to restore $\hat{\bm{A}}^{t}$ structured in the way that preserves spatial and relational coherence.
During inference, given the adjacency matrix $\bm{A}^t$ from an arbitrary frame $t$, we consider it noisy and utilize the well-trained denoising U-Net above to refine it, so as to deliver an updated version that indicates the interaction between objects, as well as the location and category of objects: $\bm{A}^{t,k-1}=\frac{1}{\sqrt[]{\alpha_t}}(\bm{A}^{t,k}-\frac{\beta_t}{\sqrt{1-\overline{\alpha}}}\epsilon_\theta(\bm{A}^{t,k},k).$
Unlike the training process which utilizes $\hat{\bm{A}}$ derived from ground truth bounding boxes,  ${\bm{A}}$ here  is obtained from the predicated ones. 
Note that the main challenge of VSGG lies in modeling of the dynamic changes of relation between object and subject, and the diffusion models primarily focus on recovering object features and their relationships. Thus, the slight bias in predicated bounding boxes, especially after ROIAlign, causes negligible impacts to relation predictions. 

\begin{figure}[t]
    \begin{center}
    \includegraphics[width=1\linewidth]{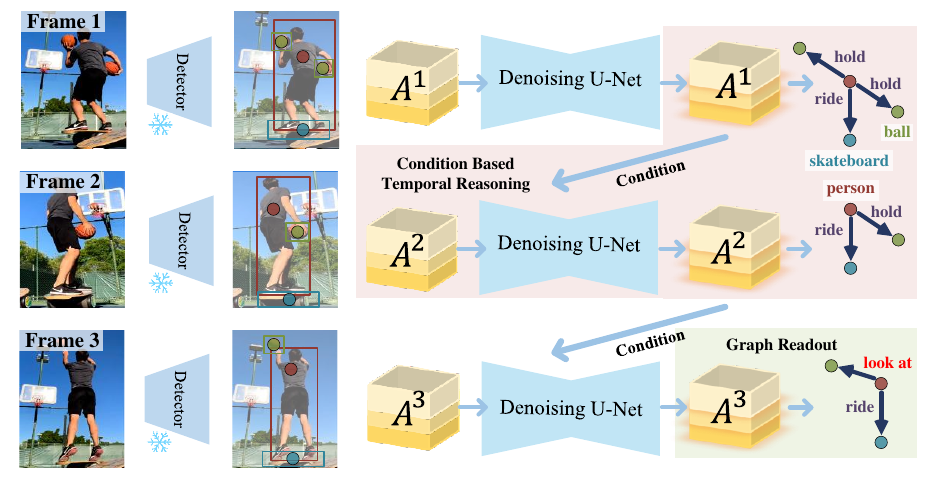}
    \end{center}
    \vspace{-18pt}
  \captionsetup{font=small}
\caption{\small{Overview of our proposed temporal prompting strategy.}
}
      \label{fig:3} 
       \vspace{-3pt}
  \end{figure}

\noindent\textbf{Condition Based Temporal Reasoning.} 
For a given video stream, we can infer future frames based on the content of preceding frames. This observation  serves as the motivation for our temporal prompting strategy (\ie, Fig.\!~\ref{fig:3}), where we condition the denoising process of $\bm{A}^{t}$ on the denoised adjacency matrix of prior frames (\ie, $\bm{A}^{t-1,0}$), as follows:
\vspace{-3pt} 
\begin{equation} 
\begin{aligned}
\bm{A}^{t,k-1}\!=\!\frac{1}{\sqrt[]{\alpha_t}}(\bm{A}^{t,k}\!-\!\frac{\beta_t}{\sqrt{1-\overline{\alpha}}}\epsilon_\theta(\bm{A}^{t,k},k,\bm{A}^{t-1,0}),
\end{aligned}\label{eq:removenoise2}
  \vspace{-2pt} 
\end{equation}
where the condition decoder $c_\theta$ in Eq.\!~\ref{eq:ldmloss} is discarded, as $\bm{A}^{t-1,0}$ has already shared the same dimension as $\bm{A}^{t,k}$. With this condition-based temporal association for denoising, the training objective for $\epsilon_\theta$ is updated as:
\begin{equation}
\mathcal{L}_\text{T\_VSGG}\!:=\!\mathbb{E}_{\bm{A}^{t,k}, \bm{A}^{t-1,0}, \epsilon,  k}\Big[ \Vert \epsilon - \epsilon_\theta(\bm{A}^{t,k}, \bm{A}^{t-1,0}, k) \Vert_{2}^{2}\Big].
\label{eq:temporalloss}
\end{equation}
Here $\mathcal{L}_\text{T\_VSGG}$ encourages $\epsilon_\theta$ to generate temporally consistent outputs, enabling a smooth and coherent denoising of adjacency matrix across frames.
Moreover, since new instances keep emerging as the video progresses, we calculate similarities between ROIAligned features of the potential new object (\ie, $\bm{F}^t_{o_\text{p}}$)  and all other objects in the past frame (\ie, $\bm{F}^{t-1}_{o_i}$), and consider $o_p$ as  new object if the similarities is smaller than 0.2 for all other objects. Then,
 the random padded features in $\bm{A}$ during graph construction are replaced with union and instance features of these new instances.

\noindent\textbf{Motion Enhanced Denoising for VSGG.} Motion cues serve as indicators to describe object movements in a given scene. They provide essential context for understanding the position and duration of events. Moreover, the speed and direction of object movements can reveal intentions for interactions (\eg, approaching), which aids in temporal reasoning by inferring whether they might complete an action or reach a goal. Therefore, given a box prediction $b^t_i=\{x_i^t,$ $y_i^t, w_i^t, h_i^t\}$ for object $i$,
we explicitly calculate the approaching speed between object $i$ and $j$ 
as follows:
\vspace{-3pt} 
\begin{equation} 
\begin{aligned}
d_{i,j}^t&=\sqrt{(x_i^t-x_j^{t})^2\!+\!(y_i^t-y_j^{t})^2},\\ v_{i,j}^t&=(d_{i,j}^t-d_{i,j}^{t-\Delta t})/{\Delta t},
\end{aligned}\label{eq:removenoise2}
  \vspace{-2pt} 
\end{equation}
 where $\Delta t$ is the interval of frame. Given $v^t\in \mathbb{R}^{N^t\times N^t}$, we pad it into the same size with $\bm{A}^t$ (\ie, $v^t\in\mathbb{R}^{N\times N}$) and inject it into $\bm{A}^{t,k}$ at each denoising step $k$: $\bm{A}^{t,k} = \bm{A}^{t,k}+v^t$.

\noindent\textbf{Graph Readout.} Next we investigate how to deliver SGG predictions for each frame based on the denoised adjacency matrix ${\bm{A}^{t,0}}$. Specifically, the prediction for predicate of each subject-object pair is obtained as follows:
\begin{equation}
r_{i,j}^t = \text{softmax}(\mathcal{F}_\text{pred}({\bm{A}^{t,0}_{i,j}}))\in\mathbb{R}^{N_\text{pred\_cls}}
\label{eq:predicate}
\end{equation}
where $\mathcal{F}_\text{pred}$ is an MLP-based classifier and $N_\text{pred\_cls}$ represents the number of predicate classes. For object classification and bounding box regression, to render a global view, we utilize the entire row $i$ of  $\bm{A}^{t,0}$ (\ie, $\bm{A}^{t,0}_i$) where all elements consider $i$ as the subject, to deliver the prediction: 
\begin{align}
o_{i}^t &= \text{softmax}(\mathcal{F}_\text{obj}({\bm{A}^{t,0}_{i}}))\in\mathbb{R}^{N_\text{obj\_cls}},\label{eq:object_cls}\\ 
b_{i}^t &=\text{sigmoid}(\mathcal{F}_\text{box}({\bm{A}^{t,0}_{i}}))\in\mathbb{R}^{4},
\label{eq:box_reg}
\end{align}
where $N_\text{obj\_cls}$ is the number of object classes, $\mathcal{F}_\text{obj}$ and $\mathcal{F}_\text{box}$ are the MLP-based classifier and projector, respectively.

\subsection{Implementation Details}\label{sec:implementation}
\noindent\textbf{Network$_{\!}$ Configuration.}$_{\!}$ 
 \textsc{DiffVsgg} is an online VSGG framework built upon the iterative diffusion diagram. It comprises three components: one off-the-shelf detector to deliver object detection results, the LDMs for inter-object relationship denosing, and projector heads for object classification, predicate prediction, and bounding box regression.

\begin{table*}
  \centering
    \small
    \resizebox{1.0\textwidth}{!}{
        \setlength\tabcolsep{0.6pt}
        \renewcommand\arraystretch{1.00}
  \begin{tabular}{r|cccccc|cccccc|cccccc}
      \toprule
      \multirow{2}*{Method} & \multicolumn{6}{c|}{PredCLS} & \multicolumn{6}{c|}{SGCLS} & \multicolumn{6}{c}{SGDET} \\
      \cmidrule(lr){2-19} 
      & R@10 & R@20 & R@50 & mR@10 & mR@20 & mR@50 & R@10 & R@20 & R@50 & mR@10 & mR@20 & mR@50 & R@10 & R@20 & R@50 & mR@10 & mR@20 & mR@50 \\
      \midrule
      RelDN~\cite{zhang2019graphical}  & 20.3 & 20.3 & 20.3 & 6.2 & 6.2 & 6.2 & 11.0 & 11.0 & 11.0 & 3.4 & 3.4 & 3.4 & 9.1 & 9.1 & 9.1 & 3.3 & 3.3 & 3.3 \\
      TRACE~\cite{teng2021target} & 27.5 & 27.5 & 27.5 & 15.2& 15.2& 15.2& 14.8 &14.8 &14.8 & 8.9 &8.9 &8.9 & 13.9 & 14.5 & 14.5 & 8.2 & 8.2 & 8.2\\
      VRD~\cite{lu2016visual} & 51.7 & 54.7 & 54.7 & - & - & - & 32.4 & 33.3 & 33.3 & - & - & - & 19.2 & 24.5 & 26.0 & - & - & - \\
      Motif Freq~\cite{zellers2018neural}  & 62.4 & 65.1 & 65.1 & - & - & - & 40.8 & 41.9 & 41.9 & - & - & - & 23.7 & 31.4 & 33.3 & - & - & - \\
      MSDN~\cite{li2017scene}  & 65.5 & 68.5 & 68.5 & - & - & - & 43.9 & 45.1 & 45.1 & - & - & - & 24.1 & 32.4 & 34.5 & - & - & - \\
      VCTREE~\cite{tang2019learning}  & 66.0 & 69.3 & 69.3 & - & - & - & 44.1 & 45.3 & 45.3 & - & - & - & 24.4 & 32.6 & 34.7 & - & - & - \\
      GPS-Net~\cite{lin2020gps}  & 66.8 & 69.9 & 69.9 & - & - & - & 45.3 & 46.5 & 46.5 & - & - & - & 24.7 & 33.1 & 35.1 & - & - & - \\
      STTran~\cite{cong2021spatial}  & 68.6 & 71.8 & 71.8 & 37.8 & 40.1 & 40.2 & 46.4 & 47.5 & 47.5 & 27.2 & 28.0 & 28.0 & 25.2 & 34.1 & 37.0 & 16.6 & 20.8 & 22.2 \\
      APT~\cite{li2022dynamic}  & 69.4 & 73.8 & 73.8 & - & - & - & 47.2 & 48.9 & 48.9 & - & - & - & 26.3 & 36.1 & 38.3 & - & - & - \\
      STTran-TPI~\cite{wang2022dynamic}  & 69.7 & 72.6 & 72.6 & 37.3 & 40.6 & 40.6 & 47.2 & 48.3 & 48.3 & 28.3 & 29.3 & 29.3 & 26.2 & 34.6 & 37.4 & 15.6 & 20.2 & 21.8 \\
      TR2~\cite{wang2023cross}  & 70.9 & 73.8 & 73.8 & - & - & - & 47.7 & 48.7 & 48.7 & - & - & - & 26.8 & 35.5 & 38.3 & - & - & - \\ 
      TEMPURA~\cite{nag2023unbiased}  & 68.8 & 71.5 & 71.5 & 42.9 & 46.3 & 46.3 & 47.2 & 48.3 & 48.3 & 34.0 & 35.2 & 35.2 & 28.1 & 33.4 & 34.9 & 18.5 & 22.6 & 23.7 \\
      DSG-DETR~\cite{feng2023exploiting}  & - & - & - & - & - & - & 50.8 & 52.0 & 52.0 & - & - & - & 30.3 & 34.8 & 36.1 & - & - & - \\\midrule
           \textbf{\textsc{DiffVsgg}}  & \textbf{71.9} & \textbf{74.5} & \textbf{74.5} & \textbf{48.1} & \textbf{50.2} & \textbf{50.2} & \textbf{52.5} & \textbf{53.7} & \textbf{53.7} & \textbf{37.3} & \textbf{38.4} & \textbf{38.4} & \textbf{32.8} & \textbf{39.9} & \textbf{45.5} & \textbf{20.9} & \textbf{23.6} & \textbf{26.2} \\
      \bottomrule
  \end{tabular}

    } 
        \vspace{-8pt}
  \captionsetup{font=small}
  \caption{Comparison of state-of-the-art VSGG methods on Action Genome \texttt{test}~\cite{ji2020action} under the \textbf{\textit{w} constraint} setting.}
  \label{tab:w_constraint}
   \vspace{-9pt}
\end{table*}

\noindent\textbf{Training Objective.}  
The training process of \textsc{DiffVsgg} consists of two stages. In the first stage, we pre-train the denoising U-Net $\epsilon_\theta$ in Eq.\!~\ref{eq:removenoise}, using $\hat{\bm{A}}^t$ constructed from ground truth bounding box annotations. In the second stage, we optimize the MLP classifier and projector in Eq.\!~\ref{eq:predicate}-\ref{eq:object_cls}, which generate the final VSGG predictions. 
Specifically, given the relation class number $C_r$, the relation prediction $p_r$, and ground truth $y_r$, the relation classification loss is:
\vspace{-2pt} 
\begin{equation}
\textcolor{black}{\mathcal{L}_\text{pred\_cls} = - \sum\nolimits_{i=1}^{C_r} y_{r,i} \log(p_{r,i}).}
\vspace{-3pt}
\end{equation}
Similarly, given the object class number $C_o$, object prediction $p_o$, and ground truth $y_o$, the object classification loss is:
\vspace{-2pt} 
\begin{equation}
\textcolor{black}{\mathcal{L}_\text{obj\_cls} = - \sum\nolimits_{j=1}^{C_o} y_{o,j} \log(p_{o,j}).}
\vspace{-3pt}
\end{equation}
For bounding box regression, given the ground truth bounding box $t_k$ (\textit{e.g.}, $(x, y, w, h)$) and the predicted bounding box $\hat{t}_k$, the regression loss is formulated as:
\begin{equation}
\textcolor{black}{\mathcal{L}_{\text{box\_reg}} = \text{Smooth}\_{\text{L1}}(t_k - \hat{t}_k).}
\end{equation}
The training objectives for each stage are formulated as:  
\vspace{-3pt}
\begin{equation} 
\begin{aligned}
  \text{Stage 1}:& \quad \mathcal{L} = \mathcal{L}_\text{T\_VSGG},\\
  \text{Stage 2}:& \quad \mathcal{L} = \mathcal{L}_\text{pred\_cls} + \mathcal{L}_\text{obj\_cls} + 0.5\mathcal{L}_\text{box\_reg},
\end{aligned}\label{eq:loss}
\end{equation}

\section{Experiment}\label{sec:experiments}
\subsection{Experimental setting }\label{sec:setting}
\noindent\textbf{Dataset.$_{\!}$} We evaluate \textsc{DiffVsgg} on Action Genome (AG)~\cite{ji2020action}, the largest video scene graph generation dataset comprising over 10K videos extended from the Charades dataset~\cite{sigurdsson2016hollywood}. 
This dataset includes 1,715,568 predicate instances across 25 predicate classes, spanning 234,253 video frames, and features 476,229 bounding boxes across 35 object categories. The 1,715,568 instances for 25 predicate classes are divided into attention, spatial and contacting, three different types. Following prior work~\cite{li2022dynamic, cong2021spatial}, we use the same training/testing split.

 \noindent\textbf{Training.} 
In the first training stage which optimizes the  LDMs with ground-truth bounding box annotations, we set the learning rate to $10^{-4}$ and use the Adam optimizer. The batch size is set to 2048, and the model is trained for 100 epochs. Each input clip consists of five frames sampled at random time intervals, enabling conditional temporal reasoning. The denoising U-Net $\epsilon_\theta$ remains frozen once after pre-training.
In the second training stage, the classifiers and projectors are trained for 10 epochs with a batch size of 8. We use the AdamW optimizer with a learning rate of $10^{-5}$, which decays by a factor of 5 halfway through the training.

\noindent\textbf{Evaluation Setup.}
In line with previous studies\!~\cite{lu2016visual,ji2020action,li2022dynamic,feng2023exploiting}, three standard evaluation protocols are adopted:
      \begin{itemize}
          \item \textbf{PredCLS:} 
          With oracle-provided object labels and bounding boxes as well as grounding-truth subject-object pairs, PredCLS assesses the model capability to predict predicate labels for each subject-object pair.
          \item \textbf{SGCLS:} Building on PredCLS, SGCLS requires the simultaneous prediction of both predicate labels and the associated subject-object pairs for each predicate.
          \item \textbf{SGDET:} As the most challenging task, SGDET requires generating complete scene graphs from scratch, including object detection, subject-object pair selection, and predicate classification. Detection is considered accurate if the overlap between prediction and ground truth exceeds 0.5.
      \end{itemize}
    
\noindent\textbf{Evaluation$_{\!}$ Metric.} We employ the Recall@$k$ where $k \in {10, 20, 50}$ as the evaluation metric, to measure the proportion of ground truth elements within the top-$k$ predictions. Additionally, Mean-Recall@$k$ is also adopted to ensure the evaluation is not biased toward high-frequency classes.
    The evaluation is performed under two different scenarios:
    \begin{itemize}
        \item \textbf{\textit{w} constraint}: Each subject-object pair is restricted to a maximum of one predicate.
        \item \textbf{\textit{w/o} constraints}: Each subject-object pair is allowed to have multiple predicates simultaneously.
    \end{itemize} 

\begin{table*}
  \centering
    \small
    \resizebox{1.0\textwidth}{!}{
        \setlength\tabcolsep{0.6pt}
        \renewcommand\arraystretch{1.00}
  \begin{tabular}{r|cccccc|cccccc|cccccc}
      \toprule
      \multirow{2}*{Method} & \multicolumn{6}{c|}{PredCLS} & \multicolumn{6}{c|}{SGCLS} & \multicolumn{6}{c}{SGDET} \\
      \cmidrule(lr){2-7} \cmidrule(lr){8-13} \cmidrule(lr){14-19}
      & R@10 & R@20 & R@50 & mR@10 & mR@20 & mR@50 & R@10 & R@20 & R@50 & mR@10 & mR@20 & mR@50 & R@10 & R@20 & R@50 & mR@10 & mR@20 & mR@50 \\
      \midrule
      RelDN~\cite{zhang2019graphical}  & 44.2  & 75.4 & 89.2 & 31.2 & 63.1 & 75.5 & 25.0 & 41.9 & 47.9& 18.6 & 36.9 & 42.6 & 13.6 & 23.0 & 36.6 & 7.5 & 18.8 & 33.7 \\ 
      VRD~\cite{lu2016visual} & 51.7 & 54.7 & 54.7 & - & - & - & 32.4 & 33.3 & 33.3 & - & - & - & 19.2 & 24.5 & 26.0 & - & - & - \\
      Motif Freq~\cite{zellers2018neural}  & 62.4 & 65.1 & 65.1 & - & - & - & 40.8 & 41.9 & 41.9 & - & - & - & 23.7 & 31.4 & 33.3 & - & - & - \\
      MSDN~\cite{li2017scene}  & 65.5 & 68.5 & 68.5 & - & - & - & 43.9 & 45.1 & 45.1 & - & - & - & 24.1 & 32.4 & 34.5 & - & - & - \\
      VCTREE~\cite{tang2019learning}  & 66.0 & 69.3 & 69.3 & - & - & - & 44.1 & 45.3 & 45.3 & - & - & - & 24.4 & 32.6 & 34.7 & - & - & - \\
      TRACE~\cite{teng2021target} & 72.6 & 91.6 & 96.4 & 50.9 & 73.6 & 82.7 & 37.1 & 46.7 & 50.5 & 31.9 & 42.7 & 46.3 & 26.5 & 35.6 & 45.3 & 22.8 & 31.3 & 41.8 \\
      GPS-Net~\cite{lin2020gps}  & 76.0 & 93.6 & 99.5 & - & - & - & - & - & - & - & - & - & 24.5 & 35.7 & 47.3 & - & - & - \\
      STTran~\cite{cong2021spatial}  & 77.9 & 94.2 & 99.1 & 51.4 & 67.7 & 82.7 & 54.0 & 63.7 & 66.4 & 40.7 & 50.1 & 58.5 & 24.6 & 36.2 & 48.8 &20.9 & 29.7 & 39.2 \\
      APT~\cite{li2022dynamic}  & 78.5 & 95.1  & 99.2  & - & - & - & 55.1 & 65.1 & 68.7 & - & - & - & 25.7 & 37.9 & 50.1 & - & - & - \\
      TR2~\cite{wang2023cross}  & 83.1 & 96.6 & 99.9 & - & - & - & 57.2 & 64.4 & 66.2 & - & - & - & 27.8 & 39.2 & 50.0 & - & - & - \\
      TEMPURA~\cite{nag2023unbiased}  & 80.4 & 94.2 & 99.4 &61.5 & 85.1 & 98.0 & 56.3 & 64.7 & 67.9 & 48.3 & 61.1 & 66.4 & 29.8 & 38.1 & 46.4 & 24.7 & 33.9 & 43.7 \\
      DSG-DETR~\cite{feng2023exploiting}  & - & - & - & - & - & - & 59.2 & 69.1 & 72.4 & - & - & - & 32.1 & 40.9 & 48.3 & - & - & - \\\midrule
      \textbf{\textsc{DiffVsgg}}  & \textbf{83.1} & \textbf{94.5} & 99.1 & \textbf{66.3} & \textbf{90.5} & \textbf{98.4} & \textbf{60.5} & \textbf{70.5} & \textbf{74.4} & \textbf{51.0} & \textbf{64.2} & \textbf{68.8} & \textbf{35.4} & \textbf{42.5} & \textbf{51.0} & \textbf{27.2} & \textbf{37.0} & \textbf{45.6} \\
      \bottomrule
  \end{tabular}
    } 
            \vspace{-8pt}
  \captionsetup{font=small}
  \caption{Comparison of state-of-the-art VSGG methods on Action Genome \texttt{test}~\cite{ji2020action} under the \textbf{\textit{w/o} constraint} setting.}
  \label{tab:wo_constraint}
     \vspace{-6pt}
\end{table*}

  \begin{figure*}
        \vspace{-5pt}
        \begin{center}
            \includegraphics[width=1\linewidth]{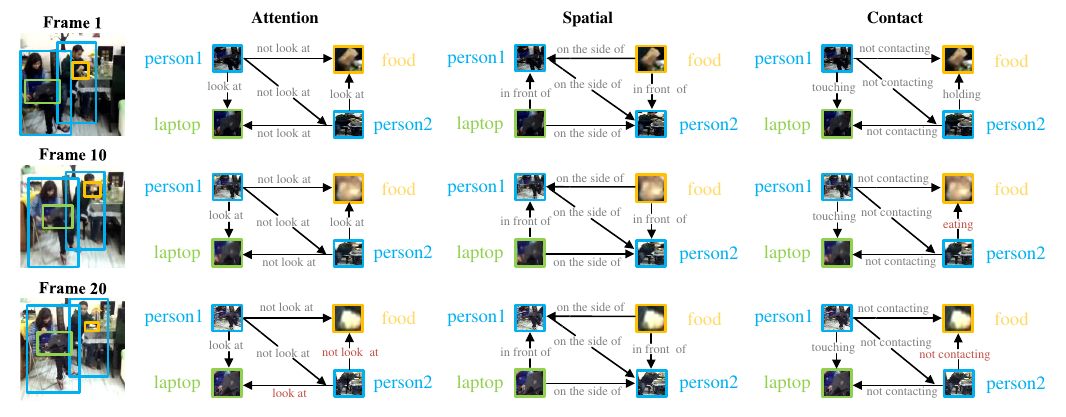}
            \end{center}
        \vspace{-18pt}
        \captionsetup{font=small}
        \caption{\small{Visualization results on Action Genome \texttt{test}\!~\cite{ji2020action}. All results are given under the \text{SGDET} setup. \textcolor{black}{Predicates in \textcolor{red}{red} indicate relationships are transformed to another one.}} See \S\ref{sec:visual} for detailed analysis.  }
        \label{fig:vis} 
        \vspace{-18pt}
    \end{figure*}

\subsection{Comparison with State-of-the-arts}\label{sec:comparison}
Tables~\ref{tab:w_constraint}-\ref{tab:wo_constraint} present the main experimental results of \textsc{DiffVsgg} against several top-leading approaches on Action Genome \texttt{test}~\cite{ji2020action}. Following existing VSGG works~\cite{li2022dynamic,cong2021spatial}, we firstly select a few representative image-level SGG methods such as RelDN~\cite{zhang2019graphical} and GPS-Net~\cite{lin2020gps}. Then, we compare \textsc{DiffVsgg}  with existing video-level solutions such as STTran\cite{cong2021spatial}, APT\cite{li2022dynamic}, \etc. In general, \textsc{DiffVsgg} outperform image-level SGG methods by a large margin in all metrics. Concretely, under the \textbf{\textit{w} constraint} setup, our proposed method achieves R@10/R@20/ R50 scores of 32.8/39.9/45.5, surpassing GPS-Net which achieves 24.7/33.1/35.1 scores by 8.1/6.8/10.4 scores, respectively. Similar trends can be observed on the \textbf{\textit{w/o} constraint} setup. All of the above firmly demonstrates the effectiveness of incorporating temporal cues to tackle relation understanding in dynamic scene.  

Additionally, when compared to the video-level counterparts, \textsc{DiffVsgg} can still deliver state-of-the-art performance. Specifically, though DSG-DETR\cite{feng2023exploiting} adopts a modern DETR-like architecture and uses a Transformer-based decoder with all frames as the inputs to aggregate temporal cues, our \textsc{DiffVsgg} which conducts online inference still outperforms it by a solid gap (\ie, 35.4/42.5/51.0 \vs 32.1/40.9/48.3 under the \textbf{\textit{w/o} constraint}). More importantly, when it comes to the comparison of mR metric which prevents the bias toward high-frequency classes, \textsc{DiffVsgg} obtains significantly higher performance compared to existing work. This suggests that using latent diffusion models (LDMs) to model the inter-object relationship is effective, and it successfully learned meaning distributions which can prevent overfitting to high-frequency classes.
Similar conclusions can be drawn on PredCLS and SGCLS two evaluation protocols, where \textsc{DiffVsgg} can still deliver SOTA performance under both \textbf{\textit{w} constraint} and \textbf{\textit{w/o} constraint} two setups.

\subsection{Qualitative Results}\label{sec:visual}
We provide visualization results of \textsc{UniAlign} in Fig.\!~\ref{fig:vis}. It can be seen that our proposed method is able to generate accurate scene graphs across various challenging scenarios, such as  fast motion and occlusion.

\subsection{Diagnostic Experiment}\label{sec:diagnostic}
To assess the effectiveness of the detailed designs of \textsc{DiffVsgg} and gain deeper insights, we conduct a series of experiments on the AG \texttt{test}\!~\cite{ji2020action}. All performance metrics are reported under the SGDET setup.

\noindent\textbf{Key Component Analysis.} We first examine the efficacy of essential components of \textsc{DiffVsgg} in Table\!~\ref{tab:design}, \textcolor{black}{where the first row denotes the baseline model directly using bounding box predictions to construct scene graphs without denoising refinement.} After integrating LDMs to model complex inter-object relationships within the scene (\ie, \textit{row} \#2), \textsc{DiffVsgg} enjoys considerable improvement on both w and w/o constraint setups. 
Next, applying condition based temporal reasoning to capture long-term temporal dependencies further boosts the performance to 32.0 and 33.9 on two setups.
Finally, upon the incorporation of motion enhanced denoising (\ie, \textit{row} \#4),  \textsc{DiffVsgg} obtains the best performance on both setups, suggesting that motion cues can effectively enhance the awareness of temporal cues through conditional promoting.

     \begin{table}
      \center
      \resizebox{0.99\columnwidth}{!}{
        \setlength\tabcolsep{1.5pt}
        \renewcommand\arraystretch{0.99}
      \begin{tabular}{c|ccc|ccc|ccc}
      \toprule
       &  & & & \multicolumn{3}{c|}{\textit{w} constraints} & \multicolumn{3}{c}{\textit{w/o} constraints} \\ \cmidrule(lr){5-7}\cmidrule(lr){8-10}
      \multirow{-2}{*}{\#} &\multirow{-2}{*}{$\mathcal{L}_\text{VSGG}$ } & \multirow{-2}{*}{$\mathcal{L}_\text{T\_VSGG}$}& \multirow{-2}{*}{Motion} & R@10 &R@20 &R@50 &R@10 &R@20 &R@50 \\
      \midrule
      1 & & & & 26.9 & 35.3 & 38.0 & 27.5 & 38.9 & 48.5 \\
      2 & \checkmark &  & & 29.7 & 37.7 & 41.5 & 30.3 & 40.1 & 49.9 \\
      3 &  &\checkmark & & 32.0 & 39.3 & 44.6 & 33.9 & 42.0 & 50.3 \\
      4 &  &\checkmark  &\checkmark & \textbf{32.8} & \textbf{39.9} & \textbf{45.5} & \textbf{35.4} & \textbf{42.5} & \textbf{51.0} \\
      \bottomrule
      \end{tabular}}
      \vspace{-3pt}
          \captionsetup{font=small}
      \caption{Analysis on key components of \textsc{DiffVsgg}.}
      \label{tab:design}
            \vspace{-4pt}
  \end{table}

   \begin{table}
      \center
      \resizebox{0.99\columnwidth}{!}{
        \setlength\tabcolsep{1.5pt}
        \renewcommand\arraystretch{0.99}
      \begin{tabular}{c|ccc|ccc|ccc}
      \toprule
       & Union Box  & Subject & Subject & \multicolumn{3}{c|}{\textit{w} constraints} & \multicolumn{3}{c}{No Constraints} \\ \cmidrule(lr){5-7}\cmidrule(lr){8-10}
      \multirow{-2}{*}{\#} &{Feature } & {Feature}& {Location} & R@10 &R@20 &R@50 &R@10 &R@20 &R@50 \\
      \midrule
      1 & \checkmark &  & & 29.5 & 36.7 & 42.0 & 32.3 & 39.2 & 47.5 \\
      2 & \checkmark &\checkmark & & 31.0 & 38.2 & 43.6 & 33.7 & 40.8 & 48.7 \\
      3 & \checkmark &\checkmark  &\checkmark & \textbf{32.8} & \textbf{39.9} & \textbf{45.5} & \textbf{35.4} & \textbf{42.5} & \textbf{51.0} \\
      \bottomrule
      \end{tabular}}
      \vspace{-3pt}
          \captionsetup{font=small}
      \caption{Analysis on elements to construct adjacency matrix $\bm{A}^t$.}
      \label{tab:adj}
            \vspace{-4pt}
  \end{table}

   \begin{table}
      \center 
      \resizebox{0.99\columnwidth}{!}{
        \setlength\tabcolsep{7.5pt}
        \renewcommand\arraystretch{0.99}
      \begin{tabular}{c|ccc|ccc}
      \toprule
       & \multicolumn{3}{c|}{\textit{w} constraints} & \multicolumn{3}{c}{\textit{w/o} constraints} \\ \cmidrule(lr){2-4}\cmidrule(lr){5-7}
      \multirow{-2}{*}{Step $T$}  & R@10 &R@20 &R@50 &R@10 &R@20 &R@50 \\
      \midrule
      10 & 31.7 & 38.5 & 42.9 & 34.0 & 41.1 & 49.5 \\
      20 & \textbf{32.8} & \textbf{39.9} & \textbf{45.5} & \textbf{35.4} & \textbf{42.5} & \textbf{51.0} \\
      50 & 33.2 & 40.5 & 46.4 & 35.7 & 43.1 & 51.4 \\
      100 &33.1 & 40.3 & 45.9 & 35.5 & 42.8 & 51.1 \\
      \bottomrule
      \end{tabular}}
      \vspace{-3pt}
          \captionsetup{font=small}
      \caption{Analysis on forward and reverse diffusion steps.}
      \label{tab:step}
            \vspace{-4pt}
  \end{table}

\noindent\textbf{Graph Construction.} 
We study the impact of using various features to construct the adjacency matrix $\bm{A}^t$ in Table~\ref{tab:adj}. As shown, incorporating all three types of features, as described in Eq.\!~\ref{eq:edge}, achieves the best performance.

\noindent\textbf{Number of Diffusion Step.} Next we investigate the effect of different number of diffusion steps. As shown in Table \ref{tab:step}, the best performance is achieved at $T=50$. However, to balance the performance and efficiency efficiency, we set $T=20$, which yields a slight reduction in performance.

\noindent\textbf{Number of Layer in Denoising U-Net.} 
We further analyze the impact of using different number of layers in the encoder and decoder of denoising U-Net. As shown in Table \ref{tab:layer}, \textsc{DiffVsgg} achieves similar performance when the layer number exceeds 2. For efficiency, we set it to 3.

\noindent\textbf{Graph Readout.} We explore different graph readout strategies in Table \ref{tab:readout}. Here $\bm{A}^{t,0}_i$ refers to utilize the entire row $i$ of $\bm{A}^{t,0}$ to deliver the predictions (\ie, Eq.\!~\ref{eq:object_cls}-\ref{eq:box_reg}), while  $\bm{A}^{t,0}_{i,j}$ denoting utilize one element at $(i,j)$ of $\bm{A}^{t,0}$ to make prediction. It can be seen that aggregating all elements in a row that contains the same subject delivers higher performance.
 
 \noindent\textbf{Running Efficiency.} Finally we probe the running efficiency of \textsc{DiffVsgg}. As seen in Table \ref{tab:fps}, our method requires the fewest trainable parameters among all competitors. Though inference speed is slightly slower than existing work due to step-wise denoising, the improvement in performance compensates for this limitation.

   \begin{table}
      \center
      \resizebox{0.99\columnwidth}{!}{
        \setlength\tabcolsep{7.5pt}
        \renewcommand\arraystretch{0.99}
      \begin{tabular}{c|ccc|ccc}
      \toprule
      \multirow{2}{*}{\# Layer} & \multicolumn{3}{c|}{\textit{w} constraints} & \multicolumn{3}{c}{\textit{w/o} constraints} \\ \cmidrule(lr){2-4} \cmidrule(lr){5-7}
      & R@10 & R@20 & R@50 & R@10 & R@20 & R@50 \\
      \midrule
        2 & 31.9 & 38.2 & 44.5 & 34.0 & 42.1 & 50.1 \\
      3 & \textbf{32.8} & \textbf{39.9} & \textbf{45.5} & \textbf{35.4} & \textbf{42.5} & \textbf{51.0} \\
      5 & 32.4 & 39.6 & 46.1 & 34.4 & 43.0 & 50.7 \\
      6 & 32.6 & 39.3 & 46.3 & 34.7 & 44.0 & 50.9 \\
      \bottomrule
      \end{tabular}}
      \vspace{-3pt}
          \captionsetup{font=small}
      \caption{Analysis of the layer configuration of the U-Net.}
      \label{tab:layer}
  \end{table}

  \begin{table}
      \center
      \resizebox{0.99\columnwidth}{!}{
        \setlength\tabcolsep{7.5pt}
        \renewcommand\arraystretch{0.99}
      \begin{tabular}{c|ccc|ccc}
      \toprule
       & \multicolumn{3}{c|}{\textit{w} constraints} & \multicolumn{3}{c}{\textit{w/o} constraints} \\ \cmidrule(lr){2-4}\cmidrule(lr){5-7}
      \multirow{-2}{*}{Element}  & R@10 &R@20 &R@50 &R@10 &R@20 &R@50 \\
      \midrule
      $\bm{A}^{t,0}_i$ & \textbf{32.8} & \textbf{39.9} & \textbf{45.5} & \textbf{35.4} & \textbf{42.5} & \textbf{51.0} \\
      $\bm{A}^{t,0}_{i,j}$ & 31.7 & 37.7 & 43.5 & 33.6 & 41.2 & 49.6 \\
      \bottomrule
      \end{tabular}}
      \vspace{-3pt}
          \captionsetup{font=small}
      \caption{Analysis on the elements used for graph readout.}
      \label{tab:readout}
      \vspace{-3pt}
  \end{table}

  \begin{table}
      \center
      \resizebox{0.99\columnwidth}{!}{
        \setlength\tabcolsep{0.5pt}
        \renewcommand\arraystretch{0.99}
      \begin{tabular}{r|cc|ccc|ccc}
      \toprule
       & Parameter&Inference& \multicolumn{3}{c|}{\textit{w} constraints} & \multicolumn{3}{c}{\textit{w/o} constraints} \\ \cmidrule(lr){4-6}\cmidrule(lr){7-9}
      \multirow{-2}{*}{Model} & \multirow{-1}{*}{Number}  & \multirow{-1}{*}{Time} & R@10 &R@20 &R@50 &R@10 &R@20 &R@50 \\
      \midrule
      TRACE~\cite{teng2021target} & 66.7 M & 8.4 FPS & 13.9 & 14.5 & 14.5 & 26.5 & 35.6 & 45.3 \\
      STTran~\cite{cong2021spatial} & 51.1 M & 10.9 FPS & 25.2 & 34.1 & 37.0 & 24.6 & 36.2 & 48.8 \\
      TEMPURA~\cite{nag2023unbiased} & 53.5 M & 9.6 FPS & 28.1 & 33.4 & 34.9 & 29.8 & 38.1 & 46.4 \\
      DSG-DETR~\cite{feng2023exploiting} & 65.5 M & 7.3 FPS & 30.3 & 34.8 & 36.1 & 32.1 & 40.9 & 48.3 \\
      \textsc{DiffVsgg} & \textbf{46.7} M & 8.7 FPS & \textbf{32.8} & \textbf{39.9} & \textbf{45.5} & \textbf{35.4} & \textbf{42.5} & \textbf{51.0} \\
      \bottomrule
      \end{tabular}}
      \vspace{-3pt}
          \captionsetup{font=small} 
      \caption{Analysis on the trainable parameters and inference time.}
      \label{tab:fps}
  \end{table}

  \section{Conclusion} 

  We presented \textsc{DiffVsgg}, a diffusion-based VSGG solution that makes contributions in three aspects. First, as an online\\ approach, it effectively addresses challenges such as  GPU memory constraints and real-time processing while delivering high performance, highlighting its potential for real-world application. Second, we propose a learning strategy for LDMs to generate inter-object relationships from random subject-object pairs, with ground-truth object locations  as the annotations. Third, in leverage of LDMs, \textsc{DiffVsgg} provides a new approach to enable temporal reasoning via dynamic, iterative refinement of scene graphs across frames, conditioned on predictions of prior frame. We hope this work could provide a new perspective for video analysis through reverse diffusion along the temporal dimension.

{
    \small
    \bibliographystyle{ieeenat_fullname}
    \bibliography{egbib}
}

\newpage
  \begin{center}
    {\Large \bfseries Supplemental Material \par}
\end{center}

\setcounter{section}{0}
\setcounter{figure}{0}
\setcounter{table}{0}

The appendix is \textbf{structured} as follows:
\begin{itemize}
	\setlength{\itemsep}{0pt}
	\setlength{\parsep}{0pt}
	\setlength{\parskip}{0pt}

	\item \S\ref{sec:s1} supplements additional implementation details.
	\item \S\ref{sec:s2} analyzes additional quantitative results.
	\item \S\ref{sec:s3} provides pseudo codes of major components of \textsc{DiffVsgg}.
	\item \S\ref{sec:s4} shows an additional diagram of the diffusion-based reasoning process.
	\item \S\ref{sec:s5} presents more visualization results.
	\item \S\ref{sec:s6} demonstrates failure case analysis.
	\item \S\ref{sec:s7} boardly discusses the limitations and social impact.
\end{itemize}

\section{More Implementation Details}~\label{sec:s1} 

\noindent\textbf{Network Configuration.} Conventional denoising diffusion models typically follow a multi-step image-to-noise process, wherein noise is progressively added and subsequently reversed to generate refined outputs. However, these models are computationally intensive. To mitigate this,\textsc{DiffVsgg} adopts an efficient forward diffusion strategy that decouples the traditional multi-step image-to-noise process into two sub-processes: image-to-zero and zero-to-noise. The projector heads are implemented as two $1\times 1$ convolutional layers with ReLU as the activation function.

\noindent\textbf{Testing.} Following established conventions\!~\cite{feng2023exploiting,wang2023cross,nag2023unbiased}, we apply no test-time data augmentation. 
Images are resized to a maximum of 720 pixels on the longer edge, with 50 steps of reverse diffusion are conducted to remove noise.

\section{Additional Quantitative Analysis}~\label{sec:s2}
\noindent\textbf{Additional Backbone.} To validate the generalization capability of \textsc{DiffVsgg}, we also conducted experiments using \textbf{ResNet-50 + DETR} as an alternative backbone. The results, shown in Table~\ref{tab:with_constraint_result}, demonstrate that \textsc{DiffVsgg} achieves state-of-the-art performance in SGDET, SGCLS and PredCLS, particularly in PredCLS, where it surpasses OED~\cite{wang2024oed} with a notable improvement of 1.8\%, 1.4\%, and 1.4\% in \textbf{R@10}, \textbf{R@20}, and \textbf{R@50} under \textbf{with constraint} setting. In addition, under \textbf{without constraint} setting, \textsc{DiffVsgg} still achieves competitive performance in terms of both mR and R metrics. In SGDET, the performance improvement of \textbf{R@10}, \textbf{R@20}, and \textbf{R@50} are 1.9\%, 1.1\%, and 1.3\%, respectively. These improvements verify the significant efficacy of \textsc{DiffVsgg} in leveraging its diffusion-based reasoning mechanism to capture object relationships and temporal dynamics across frames.

\noindent\textbf{Additional Dataset.} To further assess the adaptability of \textsc{DiffVsgg}, we conducted experiments on the \textbf{ImageNet-VidVRD} dataset as an additional benchmark. Compared to AG~\cite{ji2020action} which mostly contains human-related interaction at indoor scenes, \textbf{ImageNet-VidVRD}~\cite{shang2017video} focuses on a wider range of relations not limited to human-centric interactions, where the average number of relations and objects is 9.7 and 2.5 in each frame respectively. Videos in VidVRD are selected with the criteria of whether they have clear visual relations, containing 35 object categories and 132 relation categories, respectively. Following the standard evaluation task settings and metrics~\cite{gao2022classification,shang2021video,chen2021social}, we utilize relation tagging (RelTag) and relation detection (RelDet) to evaluate the performance of \textsc{DiffVsgg}:

\begin{itemize}[left=0pt, labelsep=1em, itemsep=1ex, topsep=0pt] 
    \item \textbf{RelTag} is the task to find all object categories and existing relations. It needs to determine whether the top-K classification results of triplets occur within the videos without considering the localization of objects and relations. We employ top-K precision (P@k , where k = 1, 5, 10) as the evaluation metric.
    \item \textbf{RelDet} is a more comprehensive task for both evaluating object categories, trajectories and existing relations. Same as AG, Recalls and mAP of relation detection are used to evaluate our model. The threshold for viewing a predicted box as a hit is 0.5.
\end{itemize}

The results, presented in Table~\ref{tab:vidvrd_results}, compare \textsc{DiffVsgg} with previous methods on the \textbf{ImageNet-VidVRD} dataset. Compared to traditional graph-based methods such as GCN~\cite{qian2019video} and STGC~\cite{liu2020beyond}, \textsc{DiffVsgg} achieves over a +10\% improvement, demonstrating its superiority in capturing contextualized information. Recent efforts such as BIG ~\cite{gao2022classification} and HCM~\cite{wei2024defense}  highlight the importance of temporal reasoning in video relation detection. Comparisons with these methods further validate the effectiveness of \textsc{DiffVsgg} in spatial-temporal reasoning modeling via denoising diffusion process. Specifically, in \textbf{RelDet} task, compared to the previous best method, HCM~\cite{wei2024defense}, \textsc{DiffVsgg} improves mAP by +0.47\%, R@50 by +0.13\%, and R@100 by +0.06\%. These improvements highlight that \textsc{DiffVsgg} is more effective in detecting and tracking relations. Similar trend is observed in the relation classification setting, \textbf{RelTag}, where \textsc{DiffVsgg} surpasses HCM by +1.45\% in P@1, +1.40\% in P@5, and +0.16\% in P@10.

\begin{table}[t]
	\centering
	\small
	\renewcommand{\arraystretch}{1.0}
	\setlength{\tabcolsep}{3pt}
	\begin{tabular}{l|ccc|ccc}
		\toprule
		\multirow{2}{*}{Method} & \multicolumn{3}{c|}{Relation Detection} & \multicolumn{3}{c}{Relation Tagging} \\ 
		\cmidrule(lr){2-4} \cmidrule(lr){5-7}
			& mAP & R@50 & R@100 & P@1 & P@5 & P@10 \\ 
		\midrule
		VidVRD~\cite{shang2017video} & 8.58 & 5.54 & 6.37 & 43.00 & 28.90 & 20.80 \\
		GSTEG~\cite{tsai2019video} & 9.52 & 7.05 & 8.67 & 51.50 & 39.50 & 28.23 \\
		MHRA~\cite{di2019multiple} & 13.27 & 6.82 & 7.39 & 41.00 & 28.70 & 20.95 \\
		GCN~\cite{qian2019video} & 14.27 & 7.43 & 8.75 & 59.50 & 40.50 & 27.58 \\
		STGC ~\cite{liu2020beyond} & 18.38 & 11.21 & 13.69 & 60.00 & 43.10 & 32.24 \\
		Fabric ~\cite{chen2021social} & 19.23 & 12.74 & 16.19 & 57.50 & 43.40 & 31.90 \\
		VidVRD-II ~\cite{shang2021video} & 23.85 & 9.74 & 10.86 & 73.00 & 53.20 & 39.75 \\
		BIG ~\cite{gao2022classification} & 26.08 & 14.10 & 16.25 & 73.00 & 55.10 & 40.00 \\
		HCM~\cite{wei2024defense} & 29.68 & 17.97 & 21.45 & 78.50 & 57.40 & 43.55 \\ 
		\hline  
		\textbf{Ours} & \textbf{30.15} & \textbf{18.10} & \textbf{21.51} & \textbf{79.95} & \textbf{58.80} & \textbf{43.71} \\ 
		\bottomrule
	\end{tabular}
	\vspace{-2mm}
	\caption{Comparison of state-of-the-art VRD methods on ImageNet-VidVRD~\cite{shang2017video}.}
	\label{tab:vidvrd_results}
\end{table}

\begin{table*}
	\centering
	  \small
	  \resizebox{1.0\textwidth}{!}{
		  \setlength\tabcolsep{1.2pt}
		  \renewcommand\arraystretch{1.0}
	\begin{tabular}{r|cccccc|cccccc|cccccc}
		\toprule
		\multirow{2}*{Method} & \multicolumn{6}{c|}{PredCLS} & \multicolumn{6}{c|}{SGCLS} & \multicolumn{6}{c}{SGDET} \\
		\cmidrule(lr){2-7} \cmidrule(lr){8-13} \cmidrule(lr){14-19}
		& R@10 & R@20 & R@50 & mR@10 & mR@20 & mR@50 & R@10 & R@20 & R@50 & mR@10 & mR@20 & mR@50 & R@10 & R@20 & R@50 & mR@10 & mR@20 & mR@50 \\
		\hline \hline
		\multicolumn{3}{c}{\textit{ResNet-101+Faster-RCNN}}\\\hline
		RelDN~\cite{zhang2019graphical}  & 20.3 & 20.3 & 20.3 & 6.2 & 6.2 & 6.2 & 11.0 & 11.0 & 11.0 & 3.4 & 3.4 & 3.4 & 9.1 & 9.1 & 9.1 & 3.3 & 3.3 & 3.3 \\
		TRACE~\cite{teng2021target} & 27.5 & 27.5 & 27.5 & 15.2& 15.2& 15.2& 14.8 &14.8 &14.8 & 8.9 &8.9 &8.9 & 13.9 & 14.5 & 14.5 & 8.2 & 8.2 & 8.2\\
		VRD~\cite{lu2016visual} & 51.7 & 54.7 & 54.7 & - & - & - & 32.4 & 33.3 & 33.3 & - & - & - & 19.2 & 24.5 & 26.0 & - & - & - \\
		Motif Freq~\cite{zellers2018neural}  & 62.4 & 65.1 & 65.1 & - & - & - & 40.8 & 41.9 & 41.9 & - & - & - & 23.7 & 31.4 & 33.3 & - & - & - \\
		MSDN~\cite{li2017scene}  & 65.5 & 68.5 & 68.5 & - & - & - & 43.9 & 45.1 & 45.1 & - & - & - & 24.1 & 32.4 & 34.5 & - & - & - \\
		VCTREE~\cite{tang2019learning}  & 66.0 & 69.3 & 69.3 & - & - & - & 44.1 & 45.3 & 45.3 & - & - & - & 24.4 & 32.6 & 34.7 & - & - & - \\
		GPS-Net~\cite{lin2020gps}  & 66.8 & 69.9 & 69.9 & - & - & - & 45.3 & 46.5 & 46.5 & - & - & - & 24.7 & 33.1 & 35.1 & - & - & - \\
		STTran~\cite{cong2021spatial}  & 68.6 & 71.8 & 71.8 & 37.8 & 40.1 & 40.2 & 46.4 & 47.5 & 47.5 & 27.2 & 28.0 & 28.0 & 25.2 & 34.1 & 37.0 & 16.6 & 20.8 & 22.2 \\
		APT~\cite{li2022dynamic}  & 69.4 & 73.8 & 73.8 & - & - & - & 47.2 & 48.9 & 48.9 & - & - & - & 26.3 & 36.1 & 38.3 & - & - & - \\
		STTran-TPI~\cite{wang2022dynamic}  & 69.7 & 72.6 & 72.6 & 37.3 & 40.6 & 40.6 & 47.2 & 48.3 & 48.3 & 28.3 & 29.3 & 29.3 & 26.2 & 34.6 & 37.4 & 15.6 & 20.2 & 21.8 \\
		TR2~\cite{wang2023cross}  & 70.9 & 73.8 & 73.8 & - & - & - & 47.7 & 48.7 & 48.7 & - & - & - & 26.8 & 35.5 & 38.3 & - & - & - \\ 
		TEMPURA~\cite{nag2023unbiased}  & 68.8 & 71.5 & 71.5 & 42.9 & 46.3 & 46.3 & 47.2 & 48.3 & 48.3 & 34.0 & 35.2 & 35.2 & 28.1 & 33.4 & 34.9 & 18.5 & 22.6 & 23.7 \\
		DSG-DETR~\cite{feng2023exploiting}  & - & - & - & - & - & - & 50.8 & 52.0 & 52.0 & - & - & - & 30.3 & 34.8 & 36.1 & - & - & - \\\hline
		\textbf{\textsc{DiffVsgg}}  & 71.9 & 74.5 & 74.5 & 48.1 & 50.2 & 50.2 & 52.5 & 53.7 & 53.7 & 37.3 & 38.4 & 38.4 & 32.8 & 39.9 & 45.5 & 20.9 & 23.6 & 26.2 \\\hline
		 \multicolumn{2}{c}{\textit{ResNet-50+DETR}}\\\hline
		OED~\cite{wang2024oed}  & 73.0 & 76.1 & 76.1 & - & - & - & - & - & - & - & - & - & 33.5 & 40.9 & 48.9 & - & - & - \\
		\hline 
		\textbf{\textsc{DiffVsgg}}  & \textbf{74.8} & \textbf{77.5} & \textbf{77.5} & \textbf{53.3} & \textbf{56.1} & \textbf{56.1} & \textbf{54.0} & \textbf{54.9} & \textbf{54.9} & \textbf{40.7} & \textbf{42.5} & \textbf{42.5} & \textbf{34.7} & \textbf{41.9} & \textbf{47.3} & \textbf{24.7} & \textbf{27.3} & \textbf{28.4} \\
		\bottomrule
	\end{tabular}
	  } 
		  \vspace{-8pt}
	\captionsetup{font=small}
	\caption{Comparison of state-of-the-art VSGG methods on Action Genome \texttt{test}~\cite{ji2020action} under the \textbf{\textit{w} constraint} setting.}
	\label{tab:with_constraint_result}
  \end{table*}

  \begin{table*}
	\centering
	  \small
	  \resizebox{1.0\textwidth}{!}{
		  \setlength\tabcolsep{1.2pt}
		  \renewcommand\arraystretch{1.0}
	\begin{tabular}{r|cccccc|cccccc|cccccc}
		\toprule
		\multirow{2}*{Method} & \multicolumn{6}{c|}{PredCLS} & \multicolumn{6}{c|}{SGCLS} & \multicolumn{6}{c}{SGDET} \\
		\cmidrule(lr){2-7} \cmidrule(lr){8-13} \cmidrule(lr){14-19}
		& R@10 & R@20 & R@50 & mR@10 & mR@20 & mR@50 & R@10 & R@20 & R@50 & mR@10 & mR@20 & mR@50 & R@10 & R@20 & R@50 & mR@10 & mR@20 & mR@50 \\
		\hline \hline
		\multicolumn{3}{c}{\textit{ResNet-101+Faster-RCNN}}\\\hline
		RelDN~\cite{zhang2019graphical}  & 44.2  & 75.4 & 89.2 & 31.2 & 63.1 & 75.5 & 25.0 & 41.9 & 47.9& 18.6 & 36.9 & 42.6 & 13.6 & 23.0 & 36.6 & 7.5 & 18.8 & 33.7 \\
		VRD~\cite{lu2016visual} & 51.7 & 54.7 & 54.7 & - & - & - & 32.4 & 33.3 & 33.3 & - & - & - & 19.2 & 24.5 & 26.0 & - & - & - \\
		Motif Freq~\cite{zellers2018neural}  & 62.4 & 65.1 & 65.1 & - & - & - & 40.8 & 41.9 & 41.9 & - & - & - & 23.7 & 31.4 & 33.3 & - & - & - \\
		MSDN~\cite{li2017scene}  & 65.5 & 68.5 & 68.5 & - & - & - & 43.9 & 45.1 & 45.1 & - & - & - & 24.1 & 32.4 & 34.5 & - & - & - \\
		VCTREE~\cite{tang2019learning}  & 66.0 & 69.3 & 69.3 & - & - & - & 44.1 & 45.3 & 45.3 & - & - & - & 24.4 & 32.6 & 34.7 & - & - & - \\
		TRACE~\cite{teng2021target} & 72.6 & 91.6 & 96.4 & 50.9 & 73.6 & 82.7 & 37.1 & 46.7 & 50.5 & 31.9 & 42.7 & 46.3 & 26.5 & 35.6 & 45.3 & 22.8 & 31.3 & 41.8 \\
		GPS-Net~\cite{lin2020gps}  & 76.0 & 93.6 & 99.5 & - & - & - & - & - & - & - & - & - & 24.5 & 35.7 & 47.3 & - & - & - \\
		STTran~\cite{cong2021spatial}  & 77.9 & 94.2 & 99.1 & 51.4 & 67.7 & 82.7 & 54.0 & 63.7 & 66.4 & 40.7 & 50.1 & 58.5 & 24.6 & 36.2 & 48.8 &20.9 & 29.7 & 39.2 \\
		APT~\cite{li2022dynamic}  & 78.5 & 95.1  & 99.2  & - & - & - & 55.1 & 65.1 & 68.7 & - & - & - & 25.7 & 37.9 & 50.1 & - & - & - \\
		TR2~\cite{wang2023cross}  & 83.1 & 96.6 & \textbf{99.9} & - & - & - & 57.2 & 64.4 & 66.2 & - & - & - & 27.8 & 39.2 & 50.0 & - & - & - \\
		TEMPURA~\cite{nag2023unbiased}  & 80.4 & 94.2 & 99.4 &61.5 & 85.1 & 98.0 & 56.3 & 64.7 & 67.9 & 48.3 & 61.1 & 66.4 & 29.8 & 38.1 & 46.4 & 24.7 & 33.9 & 43.7 \\
		DSG-DETR~\cite{feng2023exploiting}  & - & - & - & - & - & - & 59.2 & 69.1 & 72.4 & - & - & - & 32.1 & 40.9 & 48.3 & - & - & - \\\hline
  
		\textbf{\textsc{DiffVsgg}}  & 83.1 & 94.5 & 99.1 & 66.3 & 90.5 & 98.4 & 60.5 & 70.5 & 74.4 & 51.0 & 64.2 & 68.8 & 35.4 & 42.5 & 51.0 & 27.2 & 37.0 & 45.6 \\
		\hline
		 \multicolumn{2}{c}{\textit{ResNet-50+DETR}}\\\hline
		TPT~\cite{zhang2023end}  & \textbf{85.6} & \textbf{97.4} & \textbf{99.9} & - & - & - & - & - & - & - & - & - & 32.0 & 39.6 & 51.5 & - & - & - \\
		OED~\cite{wang2024oed}  & 83.3 & 95.3 & 99.2 & - & - & - & - & - & - & - & - & - & 35.3 & 44.0 & 51.8 & - & - & - \\
		\hline 
		\textbf{\textsc{DiffVsgg}}  & 84.5 & 95.9 & 99.5 & \textbf{67.9} & \textbf{91.6} & \textbf{98.9} & \textbf{62.3} & \textbf{71.8} & \textbf{75.9} & \textbf{52.7} & \textbf{65.1} & \textbf{69.5} & \textbf{37.4} & \textbf{45.1} & \textbf{53.1} & \textbf{29.7} & \textbf{37.9} & \textbf{46.4} \\
		\bottomrule
	\end{tabular}
	  } 
		  \vspace{-8pt}
	  \captionsetup{font=small}
	  \caption{Comparison of state-of-the-art VSGG methods on Action Genome \texttt{test}~\cite{ji2020action} under the \textbf{\textit{w/o} constraint} setting.}
	  \label{tab:wo_constraint}
	\end{table*}

\section{Pseudo Code}~\label{sec:s3}
We provide pytorch-style pseudo code of the proposed Graph Construction strategy in Algorithm \ref{alg:1} and Conditional Temporal Reasoning in Algorithm \ref{alg:2}.

\begin{figure*}[t]
		\begin{center}
			\includegraphics[width=\linewidth]{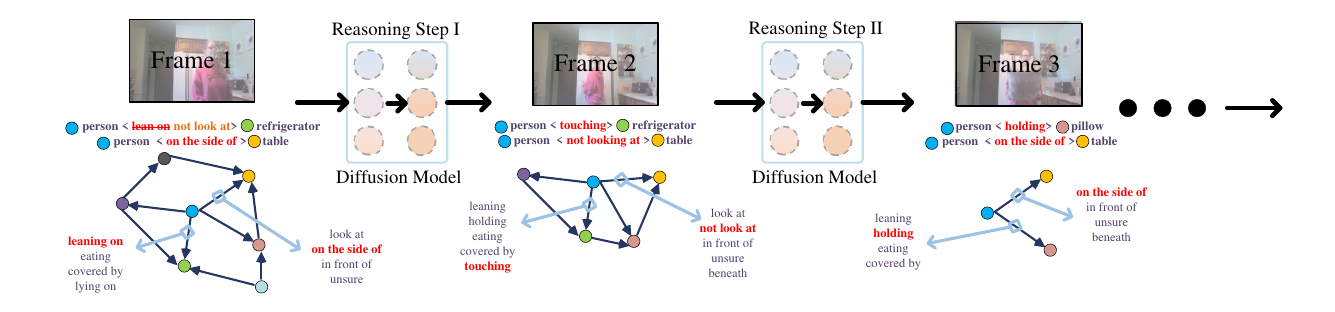}
		\end{center}
		\vspace{-20pt}
		\captionsetup{font=small}
		\caption{\small A diagram illustrating step-by-step reasoning process of \textsc{DiffVsgg}.}
		\label{fig:4}
	\end{figure*}

  \begin{algorithm}[t]
    \caption{Pseudo-code of Graph Construction in a PyTorch-like style.}
    \label{alg:1}
    \definecolor{codeblue}{rgb}{0.25,0.5,0.5}
    \lstset{
      backgroundcolor=\color{white},
      basicstyle=\fontsize{7.8pt}{7.8pt}\ttfamily\selectfont,
      columns=fullflexible,
      breaklines=true,
      captionpos=b,
      escapeinside={(:}{:)},
      commentstyle=\fontsize{7.8pt}{7.8pt}\color{codeblue},
      keywordstyle=\fontsize{7.8pt}{7.8pt},
    }
    \begin{lstlisting}[language=python]
  """
  I: input video sequence with frames I^1, I^2, ..., I^t.
  F_det: pretrained object detector.
  N_max: Max number of objects in a frame
  """
  
  # Detect
  # B_t: set of bounding boxes detected in frame t.
  # b_i : bounding box for object i.
  # b_j : bounding box for object j
  # O_t: set of class predictions for objects in frame t.
  # F_t: feature map extracted from frame t.
  F_t, b_i, b_j, O_t = (:\color{codefunc}{\textbf{F\_det}}:)((I^t))
  
  # Perform graph construction for each video frame
  (:\color{codedefine}{\textbf{def}}:) (:\color{codefunc}{\textbf{construct\_adjacency\_matrix}}:)(F_t, b_i, b_j, O_t):
     
     N_max = 100
     D_embedding = 128
     # extract instance-level feature for object i
     F_o_i = (:\color{codefunc}{\textbf{roi\_align}}:)(F_t, b_i)   
     # extract union feature from union box of i and j
     F_union = (:\color{codefunc}{\textbf{roi\_align}}:)(F_t, (:\color{codefunc}{\textbf{torch.cat}}:)([b_i, b_j], dim=1)) 
     # F_b_i box-to-feature mapping for bounding box of object j
     F_b_i = (:\color{codefunc}{\textbf{box\_to\_map}}:)(b_j) 
  
     (:\color{codedefine}{\textbf{return}}:) (:\color{codefunc}{\textbf{torch.cat}}:)([F_o_i, F_union, F_b_j], dim=-1)
  
  # Initialize adjacency matrix
  N_t = B_t.shape[0]
  A_t = (:\color{codefunc}{\textbf{torch.zeros}}:)(N_max, N_max, D_embedding)
  
  (:\color{codedefine}{\textbf{for}}:) i (:\color{codefunc}{\textbf{in range}}:)(N_t):
      (:\color{codedefine}{\textbf{for}}:) j (:\color{codefunc}{\textbf{in range}}:)(N_t):
          (:\color{codedefine}{\textbf{if}}:) i != j:
              At[i, j] = (:\color{codefunc}{\textbf{construct\_adjacency\_matrix}}:)(F_t, b_i, b_j, O_t):
    
  # Pad to fixed size
  (:\color{codedefine}{\textbf{if}}:) N_t < N_max:
     padding = (:\color{codefunc}{\textbf{torch.randn}}:)(N_max-N_t, N_max, D_embedding)
  
     A_t[N_t:, :, :] = padding
     A_t[:, N_t:, :] = (:\color{codefunc}{\textbf{torch.transpose}}:)(padding, 0, 1)
  
     (:\color{codedefine}{\textbf{return}}:) A_t
  
  \end{lstlisting}
  \end{algorithm}

  \begin{algorithm}[t]
    \caption{Pseudo-code of Temporal Reasoning in a PyTorch-like style.}
    \label{alg:2}
    \definecolor{codeblue}{rgb}{0.25,0.5,0.5}
    \lstset{
      backgroundcolor=\color{white},
      basicstyle=\fontsize{7.8pt}{7.8pt}\ttfamily\selectfont,
      columns=fullflexible,
      breaklines=true,
      captionpos=b,
      escapeinside={(:}{:)},
      commentstyle=\fontsize{7.8pt}{7.8pt}\color{codeblue},
      keywordstyle=\fontsize{7.8pt}{7.8pt},
    }
    \begin{lstlisting}[language=python]
  """
  A_t_k: Noisy adjacency matrix at step k for frame t.
  A_t_prev_0: Denoised adjacency matrix from the previous frame t-1.
  A_t_k_minus_1: adjacency matrix at the end of the k step in the denoising process for frame t
  beta_t: Noise scheduling parameter for step k.
  alpha_t: Alpha value for step k. 
  epsilon_theta: Denoising model.
  epsilon_pred: The noise predicted by the denoising model.
  """
  
  # Perform temporal reasoning for each denoising step
  (:\color{codedefine}{\textbf{def}}:) (:\color{codefunc}{\textbf{temporal\_reasoning\_step}}:)(A_t_k, A_t_prev_0, beta_t, alpha_t, epsilon_theta, k):
  
      # Predict noise with temporal conditioning
      epsilon_pred = (:\color{codefunc}{\textbf{epsilon\_theta}}:)(A_t_k, k, A_t_prev_0)
      
      # Compute reverse denoising step
      A_t_k_minus_1 = (1 / (:\color{codefunc}{\textbf{torch.sqrt}}:)(alpha_t)) * (A_t_k - (beta_t / (:\color{codefunc}{\textbf{torch.sqrt}}:)(1 - alpha_t)) * epsilon_pred)
      
      (:\color{codedefine}{\textbf{return}}:) A_t_k_minus_1
  
  # Iterate through denoising steps with temporal conditioning
  (:\color{codedefine}{\textbf{def}}:) (:\color{codefunc}{\textbf{temporal\_reasoning\_process}}:)(A_t_k, A_t_prev_0, beta_schedule, alpha_schedule, epsilon_theta, K):
  
    # Iterate from step K to 1
      (:\color{codedefine}{\textbf{for}}:) k (:\color{codefunc}{\textbf{in range}}:)(K, 0, -1):  
          A_t_k = (:\color{codefunc}{\textbf{temporal\_reasoning\_step}}:)(A_t_k, A_t_prev_0, beta_schedule[k-1], alpha_schedule[k-1], epsilon_theta, k)
      
      (:\color{codedefine}{\textbf{return}}:) A_t_k
          \end{lstlisting}
  \end{algorithm}

\section{Additional Diagram}~\label{sec:s4}
We provide an additional diagram in Fig.~\ref{fig:4} to illustrate the core concept of iterative reasoning in \textsc{DiffVsgg}. This diagram highlights the step-by-step reasoning process for scene graph generation, where the graph is progressively updated and refined through a series of iterative steps that integrate spatiotemporal cues.

\section{Further Qualitative Results}~\label{sec:s5}
In this section, we provide more qualitative comparison with existing method DSG-DETR~\cite{feng2023exploiting} on Action Genome \texttt{test}\!~\cite{ji2020action}. It could be observed in Fig.~\ref{fig:1}/ Fig.~\ref{fig:3} that \textsc{DiffVsgg} performs better in distinguish difference between hard relationship such as spatial relation: ``in front of'' \textit{vs} ``behind'', and contact relation:``lying on'' \textit{vs} ``sitting on''. 

\section{Failure Case Analysis}
\label{sec:s6} 
Due to the long-tail distribution of visual relationships in Action Genome~\cite{ji2020action}, the model struggles to accurately capture tail classes in the text, leading to biased scene graph generation. We summarize the most representative failure cases in Fig.~\ref{fig:fail}. As observed, the contact category "writing" is misclassified as "touching."

\section{Discussion}
\label{sec:s7}
\begin{figure*}[t]
		\begin{center}
			\includegraphics[width=\linewidth]{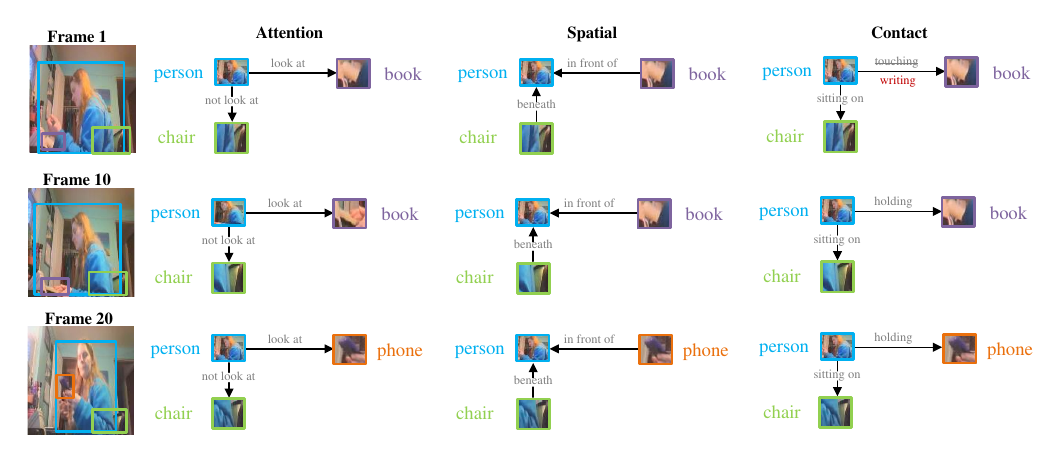}
		\end{center}
		\vspace{-28pt}
		\captionsetup{font=small}
		\caption{\small \textbf{Failure case} due to dataset bias issue.}
		\label{fig:fail}
		\vspace{-5pt}
	\end{figure*}

\noindent\textbf{Limitations.}
Although \textsc{DiffVsgg} has demonstrated remarkable performance, it has some limitations, particularly its reliance on a multi-step denoising process. Achieving high-quality outputs requires iteratively refining the predictions over numerous steps, which can be time-consuming. Drawing inspiration from recent advancements in step-reduction techniques~\cite{wimbauer2024cache, kohler2024imagine}, a potential future improvement would be to incorporate these methods to reduce the number of required steps and accelerate inference.
Another limitation of \textsc{DiffVsgg} is its dependence on the quality and distribution of training , which is also a common challenge for other VSGG methods. Biased predicate sample distributions within the dataset can lead to spurious correlations between input object pairs and predicate labels, negatively impacting the model's accuracy, especially for long-tail categories.
Fig.~\ref{fig:fail} illustrates several failure cases where biased scene graphs are generated due to the long-tailed distribution of predicates in the Action Genome dataset~\cite{ji2020action}. We aim to address this limitation by introducing additional training strategy to debias the predicate learning in our future work

\noindent\textbf{Broader Impact.}
Currently, we have only demonstrated the effectiveness of the denoising diffusion model in the VSGG task. However, the temporal reasoning capabilities across frames via diffusion offer valuable insights for designing task-specific condition prompting in related vision tasks. Integrating more informative cues from preceding frames could be an effective starting point for improving current frame predictions. The proposed continuous temporal reasoning approach could potentially be extended to tasks such as Action Recognition (AR), Video Event Detection (VED), Video-based Human-Object Interaction (V-HOI), and Multi-Object Tracking (MOT). 

On the negative side, it is important to acknowledge the risks associated with \textsc{DiffVsgg} regarding the generation of false content and data bias. The generative nature of the model during training poses the risk of creating false information about individuals, potentially damaging their reputation and privacy, and even leading to legal and ethical challenges. 
Furthermore, if the dataset used for training contains biases or imbalances, such as underrepresentation of certain races, genders, or social groups, the model's video analysis could exacerbate existing prejudices and injustices, resulting in biased and unfair decisions in real-world applications. For example, in security surveillance or crowd analysis, this bias could lead to certain groups being disproportionately monitored or wrongly accused, while others remain under-identified, ultimately affecting social equity and public trust in the technology.

\vspace{-5pt}

\clearpage

\begin{figure*}[t]
       \begin{center}
           \includegraphics[width=\linewidth]{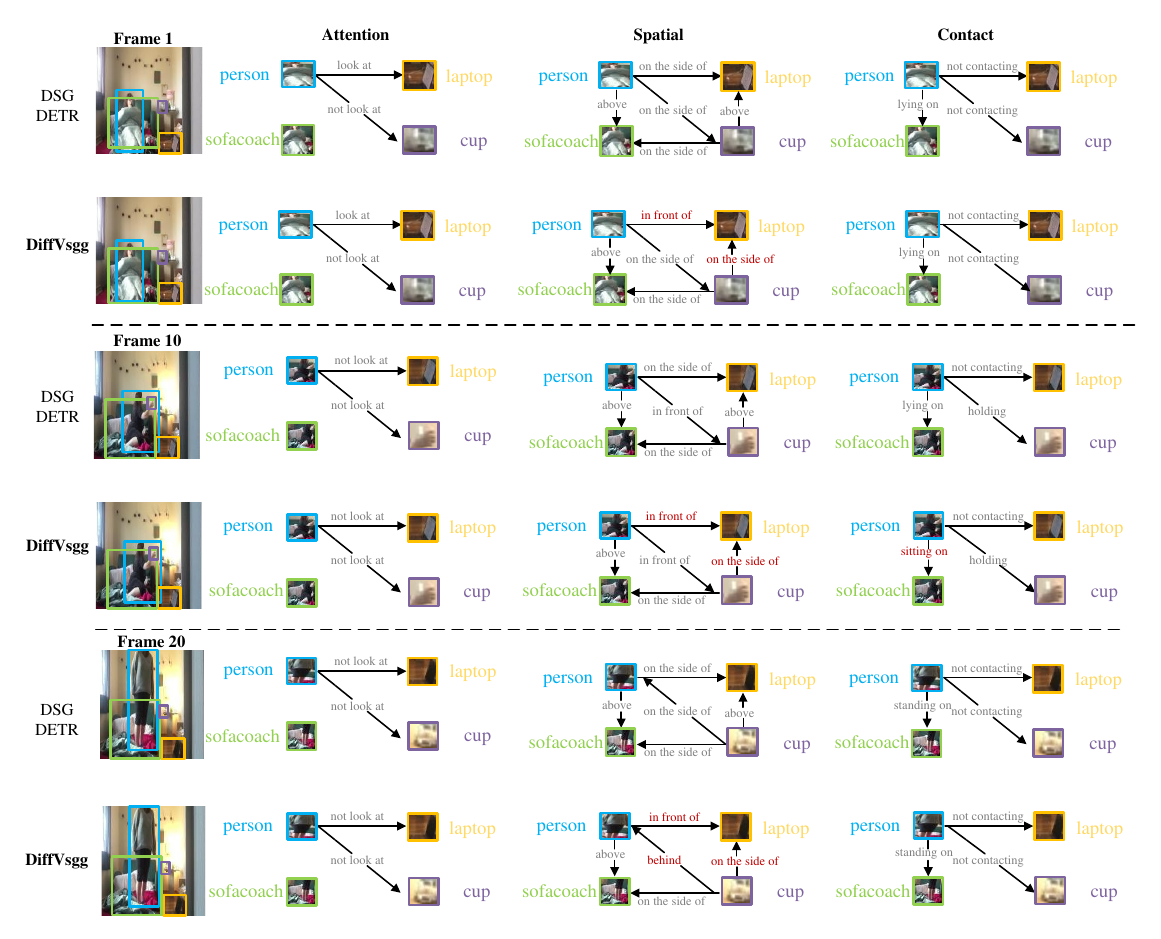}
       \end{center}
       \vspace{-25pt}
       \captionsetup{font=small}
       \caption{\small More \textbf{visual comparison} with ~\cite{feng2023exploiting} in different time steps.}
       \label{fig:1}
       \vspace{-5pt}
   \end{figure*}

   \begin{figure*}[t]
	\vspace{-245pt}
		\begin{center}
			\includegraphics[width=\linewidth]{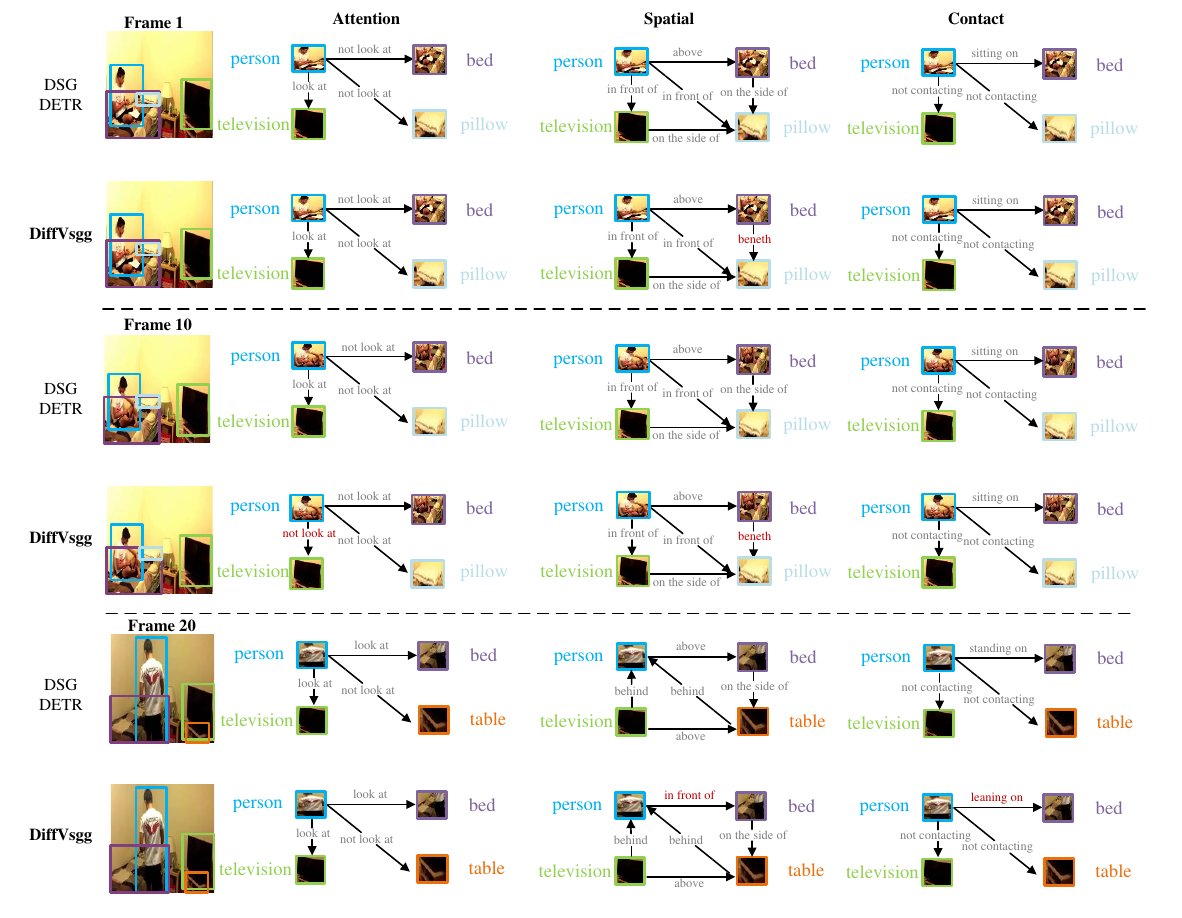}
		\end{center}
		\vspace{-27pt}
		\captionsetup{font=small}
		\caption{\small More \textbf{visual comparison} with~\cite{feng2023exploiting} in different time steps.}
		\label{fig:3}
		\vspace{-5pt}
	\end{figure*}

\end{document}